\documentclass[journal]{IEEEtran} 
\IEEEoverridecommandlockouts

\usepackage{multirow}
\usepackage{tabularx, booktabs, adjustbox}
\usepackage{balance}

\usepackage{siunitx} 
\usepackage{amssymb}

\usepackage{epsfig}
\usepackage{float}
\usepackage{color}
\usepackage{url}

\usepackage{algorithm}      
\usepackage{algorithmicx}   
\usepackage{algpseudocode}  

\algdef{SE}[DOWHILE]{Do}{doWhile}{\algorithmicdo}[1]{\algorithmicwhile\ #1}%

\DeclareMathAlphabet\mathbfcal{OMS}{cmsy}{b}{n} 

\usepackage{cite}
\usepackage{amssymb, amsfonts, bbold}
\usepackage{amsmath}
\usepackage{mathtools}
\usepackage{bbm}

\usepackage{caption}
\usepackage{subcaption}

\usepackage{bm}

\usepackage{amsthm}

\let\labelindent\relax
\usepackage{enumitem}        
\setenumerate[enumerate]{align=left}

\usepackage{multirow}

\usepackage{cite}
\makeatletter
\let\NAT@parse\undefined
\makeatother
\usepackage{hyperref}  

\newtheorem{prop}{Proposition}[section]

\newtheorem{rem}{Remark}[section]

\newcommand{\norm}[1]{\left\lVert#1\right\rVert}

\newcommand{\abs}[1]{\left\lvert#1\right\rvert}

\hypersetup{
	colorlinks=true,
	linkcolor=blue,
	filecolor=blue,      
	urlcolor=blue,
	citecolor=blue,
}

\def\R{\mathbb{R}}

\def\CovMatR{\mathbf{\Sigma}}
\def\CovMatS{\mathcal{S}}

\def\Robot{\mathcal{R}}
\def\LocPosHost{\prescript{\mathcal{L}_1}{a_1}{\mathbf{p}}} 
\def\LocPosTarg{\prescript{\mathcal{L}_2}{a_2}{\mathbf{p}}} 
\def\AlignedPosTarg{\mathbf{C} \prescript{\mathcal{L}_2}{a_2}{\mathbf{p}}} 

\def\StateVector{\mathbf{\Theta}}
\def\FIM{\mathbf{F}}
\def\detF{\det(\FIM)}
\def\detL{\det(\mathbf{\Lambda})}
\def\condF{\kappa(\hat{\FIM})}

\def\RelPosUWBAnts{\prescript{a_1}{a_2}{\mathbf{p}}}
\def\UnitZ{\mathbf{u}_z}

\DeclareMathOperator{\atantwo}{atan2}

\title{Relative Transformation Estimation Based on Fusion of Odometry and UWB Ranging Data}

\author{Thien Hoang Nguyen,~\IEEEmembership{Student Member,~IEEE}
        and Lihua Xie,~\IEEEmembership{Fellow,~IEEE}
\thanks{This work is supported by the National Research Foundation, Singapore under its Medium Sized Center for Advanced Robotics Technology Innovation. (Corresponding author: Lihua Xie.)}
\thanks{The authors are with School of Electrical and Electronic Engineering, Nanyang Technological University, Singapore 639798, 50 Nanyang Avenue (e-mail: e180071@e.ntu.edu.sg, elhxie@ntu.edu.sg).}
}

\usepackage[para]{footmisc} 

\begin{document}

\bstctlcite{BSTcontrol}

\maketitle

\begin{abstract}
In this work, the problem of 4 degree-of-freedom (3D position and heading) robot-to-robot relative frame transformation (RFT) estimation using onboard odometry and inter-robot distance measurements is studied. Firstly, we present a theoretical analysis of the problem, namely the derivation and interpretation of the Cramér-Rao Lower Bound (CRLB), the Fisher Information Matrix (FIM) and its determinant. Secondly, we propose optimization-based methods to solve the problem, including a quadratically constrained quadratic programming (QCQP) and the corresponding semidefinite programming (SDP) relaxation. Moreover, we address practical issues that are ignored in previous works, such as accounting for spatial-temporal offsets between the ultra-wideband (UWB) and odometry sensors, rejecting UWB outliers and checking for singular configurations before commencing operation. Lastly, extensive simulations and real-life experiments with aerial robots show that the proposed QCQP and SDP methods outperform state-of-the-art methods, especially in geometrically poor or large measurement noise conditions. In general, the QCQP method provides the best results at the expense of computational time, while the SDP method runs much faster and is sufficiently accurate in most cases. 
\end{abstract}


\section{Introduction} \label{sec:intro}

Many multi-robot applications such as search and rescue, environment monitoring, cooperative localization and mapping, target tracking, and entertainment have gained increasing attention in recent years \cite{queralta2020sarsurvey,shule2020mulituwbsurvey,rizk2019cooperative}. For these operations to succeed, it is important that the individual robot's pose and sensor measurements are expressed in a common reference frame. Therefore, the problem of acquiring the instantaneous relative pose or the initial RFT between the robots are of great interest. In this paper, we refer to the former problem as relative pose estimation (RPE) and the latter as relative transformation estimation (RTE), although the terms are sometimes interchangeable in the literature.

When an external reference system is accessible (e.g., from GPS and compass, a priori known map or layout of the environment, or UWB anchors with known constellation), the robots' own poses are readily available and the relative poses can be easily computed. Hence, both RPE and RTE problems are solved. However, the application will be limited to within the area within the coverage of the external system. For example, GPS would not be reliable in environments such as indoors, underground, underwater, forests etc., no prior map would be available for an unknown environment, and localization with UWB anchors is limited to only the area where the anchors are installed.

\begin{figure}[t]
\centering
	\includegraphics[width=\linewidth]{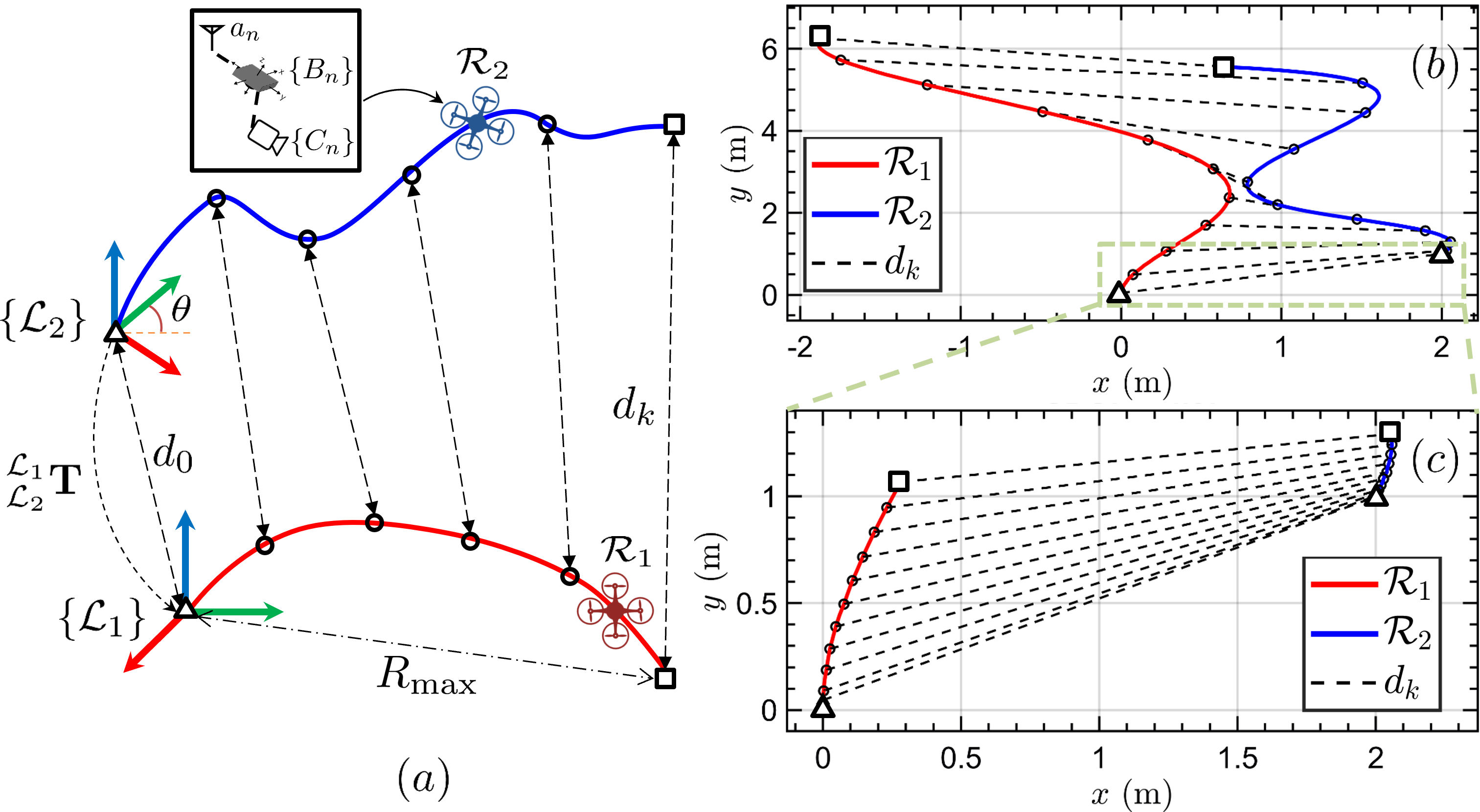}
    \caption{a) Overview of the proposed system. Our goal is to estimate $\prescript{\mathcal{L}_1}{\mathcal{L}_2}{\mathbf{T}}$. In general, the smaller the ratio $R_{\max}/d_0$ the harder the problem. b-c) Examples of “easier" (b) and “harder" (c) trajectory configurations, where (c) is the initial section of (b). Our solutions outperform previous methods in both cases.}
    \label{fig:sys_overview}
\end{figure}

Thus, many researches on using sensor measurements to obtain the relative poses have been put forth. Depending on the exteroceptive sensor (e.g., camera, LiDAR, radar, UWB) equipped onboard the robot, the inter-robot relative measurements can be obtained in the form of relative bearing, distance, position, pose, or some combinations among them \cite{zhou2013relrangebearing,knuth2013colab}. However, one implicit assumption is that the neighbor robot has known shape, size, 3D model or markers. Furthermore, the complexity and hardware cost differ greatly depending on the sensor and should be taken into consideration \cite{de2017survey}. When using camera or LiDAR, rich information regarding the neighbor robot as well as the environment can be extracted with sophisticated algorithms \cite{shenghai2021ussurvey}. In general, the performance can be limited by \cite{de2017survey}: 1) the sensor's field of view (FOV) and detecting range, 2) environmental conditions such as lighting, rain, fog, etc., 3) the complexity of the framework, which includes data processing (target detection from image/point cloud), target tracking and re-identification over time, and 4) robustness against false detection and misclassification. \textcolor{black}{In contrast, a relative localization method using only inter-robot distance measurements and local self-odometry, which is the main target of this work, can naturally overcome these challenges. Such solution would be of great appeal for many applications such as cooperative swarm \cite{cornejo2015distributed} or leader-follower formation control \cite{nguyen2019distance}.}

In this sense, UWB sensor offers several key advantages: 1) the measurement is omnidirectional with cm-level accuracy and long range (up to hundreds of meters), 2) the measurement is unaffected by lighting conditions, 3) the ranging data is comparatively simpler to model and process, 4) each UWB tag (and by extension, each robot) can be assigned a unique id without any ambiguity. Furthermore, UWB also provides a communication network \cite{nguyen2018robust} and is a more affordable, small and lightweight sensor for a team of mobile robots compared to LiDAR \cite{shenghai2021ussurvey}. On the other hand, UWB sensor might suffer under non line-of-sight (LOS) or multipath conditions and provide no information about the surrounding environment. A single ranging measurement is also not sufficiently informative for either RPE or RTE task. As a result, designing an appropriate framework that leverages the pros and alleviates the cons of UWB and existing localization systems is a promising direction that has attracted more attention recently \cite{yu2021applications}.

In this paper, our focus is on the 4 degree-of-freedom (DoF) RTE problem (i.e., estimating the 3D position and relative heading between the local frames) for a number of reasons. Firstly, if the robots' odometry data are highly accurate, the RPE and RTE problems can be seen as equivalent since the relative pose can be obtained from the initial frame transformation and the current local poses. Given that state-of-the-art simultaneous localization and mapping (SLAM) systems have achieved great progress in terms of accuracy\cite{shenghai2021ussurvey,nguyen2022viralfusion}, the RTE problem should be preferred over RPE since the state vector is much smaller. Secondly, IMU is one of the most popular sensors for SLAM systems thanks to the complementary advantages of fusing IMU and other sensors such as camera, LiDAR, radar, wheel encoder, etc. \cite{mohamed2019odomsurvey}. One of the characteristics of IMU-based SLAMs is that there are four unobservable directions \cite{huang2019visual} (three for global translation, one for rotation angle about the gravity axis). Thus, the study of the 4-DoF RTE problem would be useful to augment existing IMU-based SLAM pipelines. Thirdly, we are able to provide detailed theoretical analysis and interpretation for the 4-DoF case, which is difficult for the general 6-DoF case.

The main contributions of this work include:

\begin{itemize}

\item Theoretical analysis for the 4-DoF RTE problem and related sub-problems, including the derivation and interpretation of the CRLB, the FIM and its determinant;

\item Optimization-based approaches to solve the 4-DoF RTE problem, namely the QCQP method and its SDP relaxation, with analysis and experimental results to demonstrate the performance and tradeoff.

\item Our methods take into account practical issues, including rejecting outliers, the spatial-temporal offsets between the sensors, checking for singular configurations and providing the uncertainty of the estimates;

\item Unlike previous works where the system was often tested with only one trajectory configuration, we provide comprehensive simulation and real-life experiment results.

\end{itemize}

This paper is structured as follows. In Sect. \ref{sec:lit_review}, we review the related works on the RTE problem and highlight our contributions. The preliminaries and problem formulation are presented in Sect. \ref{sec:sys_overview}. The theoretical analysis is then presented in Sect. \ref{sec:TheorecticalAnalysis}, followed by the proposed optimization-based approaches in Sect. \ref{sec:main_approaches}. Next, simulation and experimental results in Sect. \ref{sec:exp} are used to verify the performance of our methods compared to state-of-the-arts, as well as compare the advantages and disadvantages of the proposed approaches. The concluding remarks are drawn in Sect. \ref{sec:conclusion}.

\section{Literature Review} \label{sec:lit_review}

In single-robot cases, if UWB anchors can be pre-installed in the environment, many methods have been introduced to fuse the UWB-based localization with existing onboard localization (using IMU, camera, LiDAR etc.) to reduce the error and accumulated drift \cite{yu2021applications,nguyen2020ranging}. Recent researches have relaxed the deployment requirements greatly, to even one anchor at an unknown position \cite{cao2020vir,Thien2020AuRO}.
In multi-robot cases without UWB anchors, solving the RPE task by combining UWB and vision has been studied in \cite{cao2020vir,xu2020decentralized}. However, these approaches still rely mainly on vision for detection and tracking of neighbor robots, which means the disadvantages regarding limited FOV and detecting range still apply.

If an array of UWB antennas is available on at least one robot, the range-based relative localization (RRL) problem can be solved instantaneously \cite{guler2021reloc,xianjia2021cooperative,shalaby2021relative,nguyen2018robust}. Thus, camera or LiDAR can be used for other tasks. Such systems, however, require that the robot is relatively large in size in order to accommodate the UWB antenna configuration. Alternatively, various cooperative control or motion scheduling methods have been proposed \cite{shule2020mulituwbsurvey,nguyen2019persistently,guo2019ultra,cornejo2015distributed}, which can achieve RRL at the cost of time and energy.
In our system, we assume no coordination of the robots' motion during the mission. We only require the robots' self-egomotion and a single inter-robot ranging link, which would be the most general assumptions in terms of practicality.

Algebraic and analytical solutions for the RTE problem using any combination of range and bearing measurements have been proposed in \cite{zhou2013relrangebearing,knuth2014relanycombinations}. In noisy cases, the solution might be sub-optimal and a refinement step would be necessary. As a result, a common strategy is to use a non-iterative method to find an initial solution which is then refined with nonlinear least squares (NLS) or a variant of NLS \cite{molina2019unique,li2020relSDP,jiang2020rel3D,trawny2010relplanar}. Given that any methods can be paired with an NLS refinement step, the quality of the initial guess obtained from the non-iterative method would be the main contributing factor when tested with the same dataset. \textcolor{black}{In this paper, we compare the proposed methods against previous non-iterative methods without the NLS refinement step as well as NLS with and without a good initial guess to better differentiate different approaches}.

\begin{table}[t]
\begin{adjustbox}{width=\columnwidth}
    \begin{tabular}[t]{ c|c|c}
    \toprule
        Method & Estimation problem & Approach \\
    \hline
        \multirow{2}{*}{\cite{shariati2016recovering}} & 2D translation, rotation, & Sampling-based \\
        & scale factors & convex optimization \\
    \hline
        \cite{li2020relSDP} & 2D translation, heading & SDP \\
    \hline
        \cite{molina2019unique} & 3D translation, heading & Linear \\
    \hline
        \cite{ziegler2021distributed} & 3D translation, heading & NLS \\
    \hline
        \cite{trawny2010rel3Dtransform} & 3D translation, rotation & Algebraic \\
    \hline
        \cite{jiang2020rel3D} & 3D translation, rotation & SDP \\
    \hline
        \textbf{Ours} & 3D translation, heading & QCQP, SDP \\
    \bottomrule
    \end{tabular}
\end{adjustbox}
\caption{Related works on RTE using onboard ego-motion and inter-robot range measurements. The methods \cite{molina2019unique,ziegler2021distributed,trawny2010rel3Dtransform,jiang2020rel3D} are included in our comparison.}
\label{table:lit_review}
\end{table}

The main related works are summarized in Table \ref{table:lit_review}. Among them, \cite{shariati2016recovering,li2020relSDP} only address the 2D problem, \cite{molina2019unique, ziegler2021distributed} tackle the same 4-DoF problem and \cite{jiang2020rel3D, trawny2010rel3Dtransform} solve the full 6-DoF problem.
However, the methods in \cite{molina2019unique, trawny2010rel3Dtransform} are susceptible to UWB noise, with the translation error in the order of meters when the distance standard deviation is in the order of tens of centimeters; \cite{ziegler2021distributed} relies on having a very good initial guess not too far from the true value; \cite{jiang2020rel3D} requires solving a large problem (relative to the related works) with $16$ variables and $14$ constraints while also ignoring the first distance measurement $d_0$, which is important in more challenging scenarios.
\cite{jiang2020rel3D} is the most similar to our method in the sense that an SDP problem is formulated from an original non-convex optimization problem. Lastly, there is a lack of theoretical analysis of the problem in the existing literature. Our work directly addresses all of these issues, improves the performance and provides a comparison between the SDP relaxation and the QCQP approaches which is useful for practical applications.

\section{System Overview} \label{sec:sys_overview}


\subsection{Notations} \label{subsec:notations}

Let $\prescript{A}{}{\mathbf{p}} \in \R^3$ and $\prescript{A}{}{\mathbf{R}} \in SO(3)$ be the position vector and rotation matrix in frame $\{A\}$. The corresponding quaternion of $\prescript{A}{}{\mathbf{R}}$ is $\prescript{A}{}{\mathbf{q}} \in \mathbb{H}$. $\prescript{A}{}{\mathbf{T}}$ is a homogeneous transformation matrix in frame $\{A\}$, which is defined as:
\begin{equation}
    \prescript{A}{}{\mathbf{T}} \coloneqq 
        \begin{bmatrix} 
            \prescript{A}{}{\mathbf{R}} & \prescript{A}{}{\mathbf{p}} \\ 
            \mathbf{0}^\top & 1
        \end{bmatrix}  \in SE(3).
\end{equation}
Denote $\prescript{A}{B}{\mathbf{T}}$, $\prescript{A}{B}{\mathbf{R}}$ as the transformation and rotation matrices from frame $\{B\}$ to $\{A\}$. The noisy measurement and estimated value are indicated as $(\tilde{\cdot})$ and $(\hat{\cdot})$, respectively. $\mathbb{E}[\cdot]$ is the expectation of a matrix. For simplicity, denote $\mathbf{T}_{i,j}$ as the $(i,j)$-th element of the matrix $\mathbf{T}$ and $x_i$ as the $i$-th element of vector $\mathbf{x}$. For a position vector ${\mathbf{p}} \in \R^3$, denote its elements as ${\mathbf{p}} {\coloneqq} [p^x, p^y, p^z]^\top$. Lastly, let  $t_k$ is the timestamp of the latest UWB range measurement $d_k$ and $d_0$ be the distance between the frames' origin of the two robots (Fig. \ref{fig:sys_overview}a), which is typically measured as the first distance ever received.
\subsection{Problem Formulation} \label{subsec:prob_form}

Fig. \ref{fig:sys_overview}a shows an overview of the system, which consists of a pair of robots denoted as $\Robot_n, n \in \{1,2\}$. Each robot is equipped with a UWB sensor and an IMU-based onboard odometry system. In this work, we specifically consider visual-inertial odometry (VIO) as VIO is used in our experiments. However, other IMU-based modalities \cite{shenghai2021ussurvey} still apply. Let $\{\mathcal{B}_n\}$ and $\{\mathcal{L}_n\}$ be the IMU body frame and local odometry frame, respectively. $\{\mathcal{L}_n\}$'s z-axis aligns with gravity.

During the operation, the collected data include odometry from each robot and inter-robot range measurements. At time $t_k$, the available data set is:
\begin{equation}
    \mathcal{J}_k = \{(
        \tilde{d}_i,
        \prescript{\mathcal{L}_1}{\mathcal{B}_1}{\tilde{\mathbf{p}}}_i^\top,
        \prescript{\mathcal{L}_1}{\mathcal{B}_1}{\tilde{\mathbf{q}}}_i^\top,
        \prescript{\mathcal{L}_2}{\mathcal{B}_2}{\tilde{\mathbf{p}}}_i^\top,
        \prescript{\mathcal{L}_2}{\mathcal{B}_2}{\tilde{\mathbf{q}}}_i^\top
    )\}_{i=1,...,k},
\end{equation}
where $\tilde{d}_i$ is the inter-robot UWB ranging measurement, $\prescript{\mathcal{L}_n}{\mathcal{B}_n}{\tilde{\mathbf{p}}}_i$ and $\prescript{\mathcal{L}_n}{\mathcal{B}_n}{\tilde{\mathbf{q}}}_i$ are poses of the robots in their respective local frames $\{\mathcal{L}_n\}$. 
In this work, the peer-to-peer two-way time of flight (TW-ToF) UWB ranging scheme is employed to avoid complicated clock synchronization between the sensors \cite{nguyen2020ranging}.

Without loss of generality, we assign $\Robot_1$ as the host robot and $\Robot_2$ as the target robot. Our goal is to estimate the RFT in the world frame $\{\mathcal{L}_1\}$
\begin{equation}
    \prescript{\mathcal{L}_1}{\mathcal{L}_2}{\mathbf{T}} \coloneqq
    \begin{bmatrix}
        \prescript{\mathcal{L}_1}{\mathcal{L}_2}{\mathbf{R}}& \prescript{\mathcal{L}_1}{\mathcal{L}_2}{\mathbf{p}}\\
        \mathbf{0}^\top & 1
    \end{bmatrix}
    =   
    \begin{bmatrix}
        \mathbf{C} & \mathbf{t}\\
        \mathbf{0}^\top & 1
    \end{bmatrix},
\end{equation}
with $\prescript{\mathcal{L}_1}{\mathcal{L}_2}{\mathbf{R}} \coloneqq \mathbf{C}$ and $\prescript{\mathcal{L}_1}{\mathcal{L}_2}{\mathbf{p}} \coloneqq \mathbf{t}$ for simplicity. Since it has been shown that VIO systems have four unobservable directions \cite{huang2019visual}, $\prescript{\mathcal{L}_1}{\mathcal{L}_2}{\mathbf{T}}$ can be parameterized by
\begin{equation}
    \StateVector \coloneqq [\mathbf{t}^\top, \theta]^\top = [t^x, t^y, t^z, \theta]^\top.
\end{equation}
with $\theta$ as the relative yaw angle between $\mathcal{L}_1$ and $\mathcal{L}_2$ (Fig. \ref{fig:sys_overview}a), i.e. $\mathbf{C}$ can be calculated as the basic 3D rotation matrix around the $z$-axis by an angle $\theta$ as
\begin{equation}
    \mathbf{C} =
    \begin{bmatrix}
        \cos \theta & -\sin \theta  & 0\\
        \sin \theta &  \cos \theta  & 0\\
        0           &  0            & 1
    \end{bmatrix}.
\end{equation}

\subsection{UWB measurement model}


We use the same UWB measurement model as our previous works \cite{nguyen2022viralfusion,Thien2021RAL}, which takes into account: 1) the spatial offset of the UWB antenna in the body frame, and 2) the temporal offset between UWB and odometry data.
Let $\prescript{\mathcal{B}_n}{a_n}{\mathbf{p}}$ be the position of the UWB antenna $a_n$ in the body frame $\{\mathcal{B}_n\}$, which is obtained from calibration. At time $t_k$, the UWB antenna position in the local odometry frame is:
\begin{equation}
    \prescript{\mathcal{L}_n}{a_n}{\mathbf{p}}_k =
    \prescript{\mathcal{L}_n}{B_n}{\mathbf{p}}_k + \prescript{\mathcal{L}_n}{B_n}{\mathbf{R}}_k \prescript{B_n}{a_n}{\mathbf{p}}.
\end{equation}
\textcolor{black}{
For brevity, denote ${\LocPosHost \coloneqq [\varphi^x, \varphi^y, \varphi^z]^\top}$, ${\LocPosTarg \coloneqq [o^x, o^y, o^z]^\top}$ and ${\RelPosUWBAnts \coloneqq \prescript{\mathcal{L}_1}{a_2}{\mathbf{p}} - \LocPosHost}$ as the relative position vector between the UWB antennas on the two robots in the world frame at time $t_i$.
}
If the spatial offset is negligible (i.e., $\prescript{\mathcal{B}_n}{a_n}{\mathbf{p}} \approx \mathbf{0}$), the UWB measurement model is reverted back to the one used in the literature (i.e., $d_k = \norm{\mathbf{t} + \mathbf{C} \prescript{\mathcal{L}_2}{\mathcal{B}_2}{\mathbf{p}}_k - \prescript{\mathcal{L}_1}{\mathcal{B}_1}{\mathbf{p}}_k}$). However, in some applications the UWB antenna is located at the boundary of the robot platform to ensure LOS during the operation and/or large baseline between the antennas \cite{nguyen2022viralfusion,xianjia2021cooperative}. 
In such cases, $\prescript{\mathcal{B}_n}{a_n}{\mathbf{p}}$ is not trivial and should be taken into account.

The relationship between the noisy distance measurement and the state vector can be written as
\begin{equation}
    \tilde{d}_k = d_k + \eta_k 
    = \norm{\mathbf{t} + \AlignedPosTarg_k - \LocPosHost_k} + \eta_k,
\end{equation}
where ${d_k = \norm{\mathbf{t} + \AlignedPosTarg_k - \LocPosHost_k}}$ is the true inter-robot distance. The measurement is assumed to be corrupted by independent and identically distributed Gaussian noise $\eta_k \sim \mathcal{N}(0, \sigma_{\textrm{r}}^2)$. 

\section{Theoretical Analysis} \label{sec:TheorecticalAnalysis}

In this section, we present the derivation of the FIM, the CRLB and the determinant of the FIM associated with the 4-DoF RTE problem formulated in Sect. \ref{subsec:prob_form}. These entities provide different perspectives to understand the problem, independent of any specific estimator.

\subsection{Fisher Information Matrix and Cramér-Rao Lower Bound} \label{subsec:CRLB_derivation}

Denote the noise-free and noisy distance vectors respectively as
\begin{equation}
\begin{aligned}
    \mathbf{f}(\StateVector) &= [d_1, \; d_2, \; \dots, d_k]^\top,\\
    \tilde{\mathbf{d}} &= [\tilde{d}_1, \; \tilde{d}_2, \; \dots, \tilde{d}_k]^\top,
\end{aligned}
\end{equation}
the probability density function of $\tilde{\mathbf{d}}$ is given by:
\begin{equation}
\begin{aligned}
    p(\tilde{\mathbf{d}}, \StateVector) &= \frac{1}{(2\pi)^{k/2} \sqrt{\operatorname{det}(\CovMatR)}}\\
    &. \operatorname{exp} \left[ - \frac{1}{2} (\tilde{\mathbf{d}} - \mathbf{f}(\StateVector))^\top \CovMatR^{-1} (\tilde{\mathbf{d}} - \mathbf{f}(\StateVector)) \right]
\end{aligned}
\end{equation}
where $\CovMatR = \sigma_r^2 \; \mathbf{I}_{k \times k}$.
Let $\hat{\StateVector}$ be an estimate of ${\StateVector}$. The CRLB is the lowest bound for the error covariance matrix of an unbiased estimator, i.e.
\begin{equation} \label{eq:CRLB_definition}
    \mathbb{E}\left[ 
    (\hat{\StateVector} - \StateVector) (\hat{\StateVector} - \StateVector)^\top
    \right] \geq \textrm{CRLB} = \FIM^{-1},
\end{equation}
where $\FIM$ is the FIM. The FIM encodes the amount of information provided by the set of measurements to estimate the parameters. 

In general, the $(i,j)$-th element of FIM is:
\begin{equation}
    \FIM_{i,j} \coloneqq \mathbb{E}\left[
    \frac{\partial}{\partial \Theta_i}\ln \left( p(\tilde{\mathbf{d}}, \StateVector) \right)
    \frac{\partial}{\partial \Theta_j}\ln \left( p(\tilde{\mathbf{d}}, \StateVector) \right)
    \right],
\end{equation}
where the natural logarithm of $p(\tilde{\mathbf{d}}, \StateVector)$ is \begin{equation}
    \ln \left( p(\tilde{\mathbf{d}}, \StateVector) \right) = - \frac{1}{2} (\tilde{\mathbf{d}} - \mathbf{f}(\StateVector))^\top \CovMatR^{-1} (\tilde{\mathbf{d}} - \mathbf{f}(\StateVector)) + c,
\end{equation}
with $c$ being a constant scalar. Under the i.i.d. zero-mean Gaussian noise assumption, the formulation of FIM is \cite{zekavat2011handbook}:
\begin{equation}
\begin{aligned}
    \FIM = 
    \left[ \frac{\partial \mathbf{f}(\StateVector)}{\partial \StateVector} \right]^\top 
    \CovMatR^{-1} 
    \left[ \frac{\partial \mathbf{f}(\StateVector)}{\partial \StateVector} \right]
\end{aligned}
\end{equation}
which can be rewritten as
{\color{black}
\begin{equation}
\begin{aligned}
    \FIM = 
    \frac{1}{\sigma_r^2}
    \sum\limits_{i=1}^{k}
    \mathbf{G}_i^\top \mathbf{G}_i,
\end{aligned}
\end{equation}
}
where
\begin{equation} \label{eq:G_i_orginial}
\begin{aligned}
    \mathbf{G}_i &= \left[
        \frac{\partial f_i(\StateVector)}{\partial t^x}, \;
        \frac{\partial f_i(\StateVector)}{\partial t^y}, \;
        \frac{\partial f_i(\StateVector)}{\partial t^z}, \;
        \frac{\partial f_i(\StateVector)}{\partial \theta}
    \right] \\
    &= \left[
        \partial_x f_i, \; 
        \partial_y f_i, \;
        \partial_z f_i, \;
        \partial_{\theta} f_i
    \right].
\end{aligned}
\end{equation}
The derivations of the above Jacobians can be found in Appendix \ref{appendix:FIM_full}.
The FIM can be simplified as
\begin{equation}\label{eq:FIM_simplified}
\begin{aligned}
    \FIM 
    &= \frac{1}{\sigma_r^2}
    \sum\limits_{i=1}^{k}
    \begin{bmatrix}
        (\partial_x f_i)^2 & \dots & (\partial_x f_i)(\partial_{\theta} f_i) \\
        \vdots & \ddots & \vdots \\
        (\partial_{\theta} f_i)(\partial_x f_i) & \dots & (\partial_{\theta} f_i)^2
    \end{bmatrix} \\
    &= \frac{1}{\sigma_r^2} \sum\limits_{i=1}^{k} \mathbf{G}_i^\top \mathbf{G}_i 
    = \frac{1}{\sigma_r^2} \mathbf{J}^\top \mathbf{J},
\end{aligned}
\end{equation}
where the $i$-th row of the matrix $\mathbf{J}$ is $\mathbf{G}_i$, i.e. 
$\mathbf{J} \coloneqq 
    \begin{bmatrix}
        \mathbf{G}_1\\
        \vdots\\
        \mathbf{G}_k
    \end{bmatrix}$.
From Eq. (\ref{eq:FIM_simplified}), it can be seen that the FIM: is either positive definite or semi-definite; depends on the number of available measurements ($k$); depends on the precision of the measurements ($\sigma_r$) as well as the Jacobians ($\mathbf{G}_i$). 
If the FIM is non-singular, i.e. $\detF \neq 0$, the CRLB can be obtained as the inverse of FIM ($\FIM^{-1}$).
For evaluation purposes, we additionally define the translation and heading CRLBs as
\begin{equation}
\begin{aligned}
    \textrm{CRLB}_t = \sum\limits_{i=1}^{3}  [\FIM^{-1}]_{i,i},\;\textrm{CRLB}_{\theta} = [\FIM^{-1}]_{4,4}.
\end{aligned}
\end{equation}
The $\textrm{CRLB}_t$ and $\textrm{CRLB}_{\theta}$ are used to compare the mean square error (MSE) of different methods against the theoretical bound.

\subsection{Determinant of the FIM} \label{subsec:det_FIM}
Applying the Cauchy-Binet formula to $\FIM {=} \frac{1}{\sigma_r^{2}} \mathbf{J}^\top \mathbf{J}$, we have
\begin{equation}\label{eq:det_FIM_4dof}
\begin{aligned}
    \detF
    = \frac{1}{\sigma_r^2} \sum\limits_{1 \leq j_1 < j_2 < j_3 < j_4 \leq k}^{} (\detL)^2
\end{aligned}
\end{equation}
where $\mathbf{\Lambda}$ denotes the $4\times4$ matrix consisting of the $j_1$,$j_2$,$j_3$,$j_4$-th rows of $\mathbf{J}$.
\textcolor{black}{For clarity, in the following we focus on $\mathbf{\Lambda}$ with $1{-}4$-th rows of $\mathbf{J}$, but the analysis can be generalized to any $j_1{-}j_4$-th rows.}
As shown in Appendix \ref{appendix:FIM_full}, we can write
\begin{equation} \label{eq:Lambda_def}
    \mathbf{\Lambda} \coloneqq 
    \begin{bmatrix}
        \mathbf{G}_1 \\[0.2em]
        \mathbf{G}_2 \\[0.2em]
        \mathbf{G}_3 \\[0.2em]
        \mathbf{G}_4
    \end{bmatrix} =
    \begin{bmatrix}
        \mathbf{u}_1^\top & \Phi_1 \\[0.2em]
        \mathbf{u}_2^\top & \Phi_2 \\[0.2em]
        \mathbf{u}_3^\top & \Phi_3 \\[0.2em]
        \mathbf{u}_4^\top & \Phi_4
    \end{bmatrix},
\end{equation}
where $\mathbf{u}_i {=} \frac{1}{d_i} \RelPosUWBAnts_i$ ($i \in \{1,\cdots,4\}$) is the unit vector parallel to $\RelPosUWBAnts_i$ and
\textcolor{black}{
\begin{equation} \label{eq:detF_phi_i}
    \Phi_i =
        \left( \UnitZ \times \AlignedPosTarg_i \right) 
        \cdot \mathbf{u}_i = \rho_i \sin \gamma_i,
\end{equation}
with
$\gamma_i = \frac{\pi}{2} - \measuredangle \left(\UnitZ \times \AlignedPosTarg_i, \mathbf{u}_i\right)$,
$\rho_i$ is the length of the projection of $\AlignedPosTarg_1$ on the $xy$ plane of $\{\mathcal{L}_i\}$, and
$\UnitZ = [0, 0, 1]^\top$ is the unit vector in the $z$ axis.
}

By applying the Laplace expansion along the last column of $\mathbf{\Lambda}$ in Eq. (\ref{eq:Lambda_def}), we have:
\begin{equation}\label{eq:det_Lambda_Laplace_expansion}
\begin{aligned}
    \detL = -\Phi_1 T_1 + \Phi_2 T_2 - \Phi_3 T_3 + \Phi_4 T_4
\end{aligned}
\end{equation}
where the minors are:
\begin{equation}\label{eq:det_Lambda_minors_original}
\begin{aligned}
    T_1 {=} 
    \begin{vmatrix}
        \mathbf{u}_2^\top \\[0.2em]
        \mathbf{u}_3^\top \\[0.2em]
        \mathbf{u}_4^\top
    \end{vmatrix},\;
    T_2 {=} 
    \begin{vmatrix}
        \mathbf{u}_1^\top \\[0.2em]
        \mathbf{u}_3^\top \\[0.2em]
        \mathbf{u}_4^\top
    \end{vmatrix},\;
    T_3 {=} 
    \begin{vmatrix}
        \mathbf{u}_1^\top \\[0.2em]
        \mathbf{u}_2^\top \\[0.2em]
        \mathbf{u}_4^\top
    \end{vmatrix},\;
    T_4 {=}
    \begin{vmatrix}
        \mathbf{u}_1^\top \\[0.2em]
        \mathbf{u}_2^\top \\[0.2em]
        \mathbf{u}_3^\top
    \end{vmatrix}.
\end{aligned}
\end{equation}
For any vectors $\mathbf{a}$, $\mathbf{b}$, $\mathbf{c} \in \mathbb{R}^3$, we have 
\begin{equation}
\begin{aligned}
    &\det \left(
    \begin{bmatrix}
        \mathbf{a}^\top \\
        \mathbf{b}^\top \\
        \mathbf{c}^\top
    \end{bmatrix} \right)
    = \det([\mathbf{a} \; \mathbf{b} \; \mathbf{c}]) 
    = (\mathbf{a} \times \mathbf{b}) \cdot \mathbf{c} \\
    &= \norm{\mathbf{a}} \norm{\mathbf{b}} \norm{\mathbf{c}} 
    \left| \sin \measuredangle (\mathbf{a}, \mathbf{b}) \right|
    \cos \measuredangle (\mathbf{a} \times \mathbf{b}, \mathbf{c}).
\end{aligned}
\end{equation}
\textcolor{black}{
As such, $T_1$ can be written as
\begin{equation} \label{eq:det_Lambda_T4}
\begin{aligned}
    T_1 &= 
    (\mathbf{u}_2 \times \mathbf{u}_3) 
    \cdot 
    \mathbf{u}_4 \\
    &= 
    \norm{\mathbf{u}_2} 
    \norm{\mathbf{u}_3} 
    \norm{\mathbf{u}_4} 
    \left| \sin{\measuredangle (\mathbf{u}_2, \mathbf{u}_3)} \right| 
    \cos{\measuredangle (\mathbf{u}_2 \times \mathbf{u}_3, \mathbf{u}_4)}\\
    &= \left| \sin{\alpha_1} \right| \sin{\beta_1}, \\
\end{aligned}
\end{equation}
where 
$\alpha_1 = \measuredangle (\mathbf{u}_2, \mathbf{u}_3)$,
$\beta_1 = \frac{\pi}{2} - \measuredangle (\mathbf{u}_2 \times \mathbf{u}_3, \mathbf{u}_4)$.}
Similarly, we can write:
\begin{equation} \label{eq:det_Lambda_T123}
\begin{aligned}
    T_2 &= (\mathbf{u}_1 \times \mathbf{u}_3) \cdot \mathbf{u}_4 
    = \left| \sin{\alpha_2} \right| \sin{\beta_2}, \\
    T_3 &= (\mathbf{u}_1 \times \mathbf{u}_2) \cdot \mathbf{u}_4 
    = \left| \sin{\alpha_3} \right| \sin{\beta_3}, \\
    T_4 &= (\mathbf{u}_1 \times \mathbf{u}_2) 
    \cdot \mathbf{u}_3
    = \left| \sin{\alpha_4} \right| \sin{\beta_4}.\\
\end{aligned}
\end{equation}

\begin{figure}[t]
    \begin{subfigure}[t]{\linewidth}
    \centering
    \includegraphics[width=\linewidth]{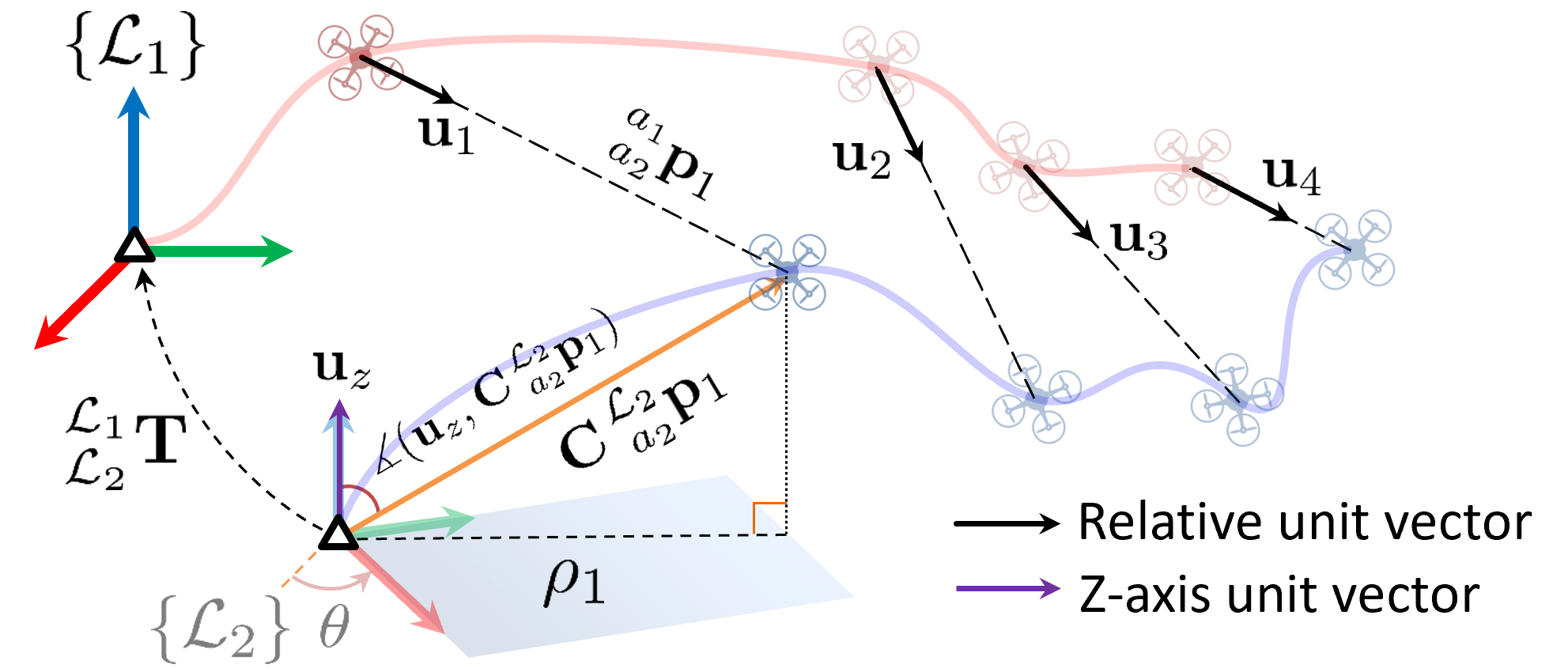}
    \caption{Illustration of the main vectors that derive $T_1$ and $\Phi_1$.}
    \label{fig:viz_vectors}
    \end{subfigure}
    \begin{subfigure}[t]{.48\linewidth}
    \centering
    \includegraphics[width=\linewidth]{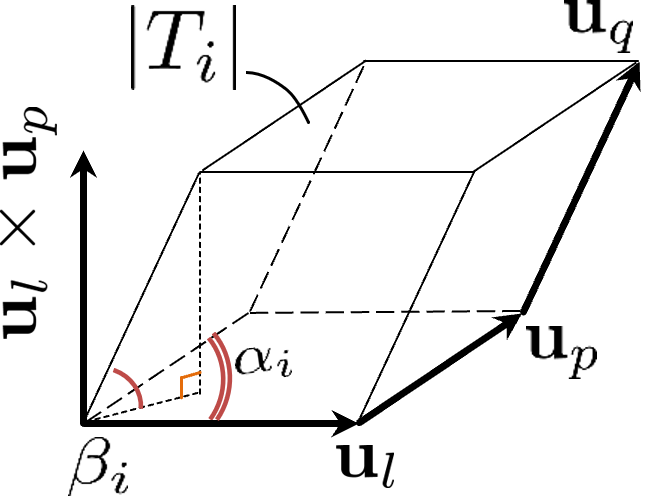}
    \caption{$T_i = 
    (\mathbf{u}_l \times  \mathbf{u}_p) 
    \cdot 
    \mathbf{u}_q$}
    \label{fig:viz_T_i}
    \end{subfigure}
    \hfill
    \begin{subfigure}[t]{.48\linewidth}
    \centering
    \includegraphics[width=\linewidth]{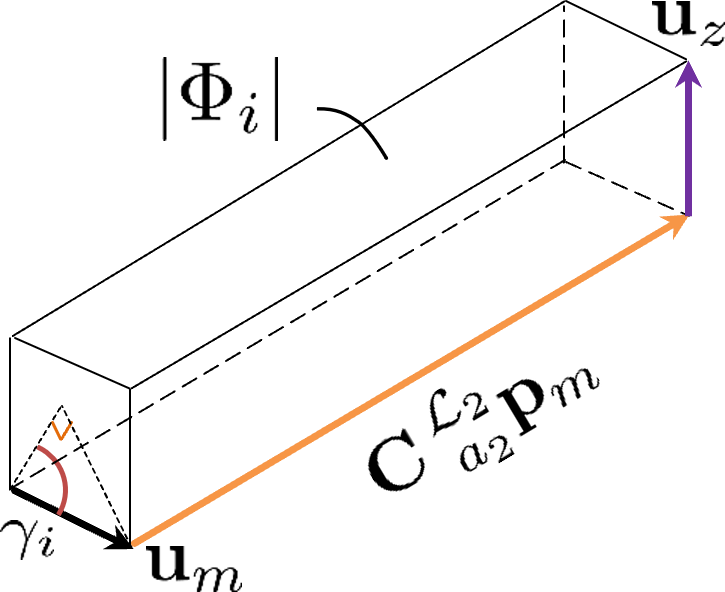}
    \caption{$\Phi_i = 
            \left( \UnitZ \times \AlignedPosTarg_m \right) 
            \cdot \mathbf{u}_m$}
    \label{fig:viz_Phi_ji}
    \end{subfigure}
    \caption{\textcolor{black}{Visualization of the vectors, the parallelepipeds defined by the three vectors $\{\mathbf{u}_l,\mathbf{u}_p,\mathbf{u}_q\}$ and  $\{\mathbf{u}_m,\AlignedPosTarg_m,\UnitZ\}$, as well as the components $\alpha_i$, $\beta_i$, $\gamma_i$ and $\rho_i$ that comprise $\detF$.}}
    \label{fig:analytical_coord_systems}
\end{figure}

Combining the above equations,
we have
\begin{equation}\label{eq:det_FIM_RTE}
\begin{aligned}
    &\detF 
    = \frac{1}{\sigma_r^2} \sum\limits_{S}^{} 
    \left[
        \sum\limits_{i=1}^{4}
        (-1)^{i} \; \Phi_i \; T_i
    \right]^2 \\
    &= \frac{1}{\sigma_r^2} \sum\limits_{S}^{} 
    \left[
        \sum\limits_{i=1}^{4}
        (-1)^{i}
        \left[
            \left( \UnitZ {\times} \AlignedPosTarg_m \right)
            \cdot 
            \mathbf{u}_m
        \right]
        \left[
            (\mathbf{u}_l \times \mathbf{u}_p) 
            \cdot 
            \mathbf{u}_q
        \right]
    \right]^2\\
    &= \frac{1}{\sigma_r^2} \sum\limits_{S}^{} 
    \left[
        \sum\limits_{i=1}^{4} (-1)^{i}
        \rho_i
        \left| \sin{\alpha_i} \right|
        \sin{\beta_i} \;
        \sin{\gamma_i}
    \right]^2
\end{aligned}
\end{equation}
where $S = \{ (j_1,j_2,j_3,j_4) \; \vert \; 1 \leq j_1 < j_2 < j_3 < j_4 \leq k \}$, ${m = j_i}$, ${(l,p,q) \in \{j_1,j_2,j_3,j_4\} \setminus m}$, ${l < p < q}$ and
\begin{equation} \label{eq:det_FIM_components}
\begin{aligned}
    &\alpha_i = 
    \measuredangle (\mathbf{u}_l, \mathbf{u}_p), \\
    &\beta_i = 
    \frac{\pi}{2} - \measuredangle (\mathbf{u}_l \times \mathbf{u}_p, \mathbf{u}_q), \\
    &\gamma_i = 
    \frac{\pi}{2} - \measuredangle 
    \left(
    \UnitZ \times \AlignedPosTarg_m, \mathbf{u}_m
    \right), \\
    &\rho_i =
    \norm{\AlignedPosTarg_m}
    \left| 
        \sin \measuredangle (\UnitZ, \AlignedPosTarg_m)
    \right|.
\end{aligned}
\end{equation}
Fig. \ref{fig:analytical_coord_systems} illustrates the components that build up $\detF$.
Note that $\beta_i$ is defined only when $\mathbf{u}_l \times \mathbf{u}_p \neq \mathbf{0}$, i.e. $\mathbf{u}_l$ and $\mathbf{u}_p$ are non-parallel. Otherwise, if $\mathbf{u}_l \times \mathbf{u}_p = \mathbf{0}$ then $T_i = 0$ and $\beta_i$ is not relevant.
Similarly, $\gamma_i$ is defined only when $\UnitZ \times \AlignedPosTarg_i \neq \mathbf{0}$, i.e. $\UnitZ$ and $\AlignedPosTarg_i$ are non-parallel, otherwise $\Phi_i = 0$ and $\gamma_i$ is irrelevant.
In the next section, we will discuss in more details the physical meaning of $\alpha$, $\beta$, $\gamma$ and $\rho$.

\begin{prop} \label{prop:sub_prob}
From the general problem of 3D RTE with unknown $\theta$ (i.e., without a common heading reference between the robots), one can derive the sub-problems, the corresponding state vector and determinant of the FIM as
\begin{enumerate}[label=(\arabic*)]

    \item 3D RTE with known $\theta$, $\StateVector \coloneqq [t^x, t^y, t^z]^\top$,
    \begin{equation}
    \begin{aligned}
        &\det(\FIM) = 
        \frac{1}{\sigma_r^2} 
        \sum\limits_{1 \leq i < j < l \leq k}^{}
        \sin^2{\alpha} 
        \sin^2{\beta},
    \end{aligned}
    \end{equation}
    where $\alpha = \measuredangle (\mathbf{u}_i, \mathbf{u}_j)$, $\beta= \frac{\pi}{2} - \measuredangle (\mathbf{u}_i \times \mathbf{u}_j, \mathbf{u}_l)$.
    \item 2D RTE with unknown $\theta$, $\StateVector \coloneqq [t^x, t^y, \theta]^\top$,
    \begin{equation}
    \begin{aligned}
        &\det(\FIM) = 
        \frac{1}{\sigma_r^2} \sum\limits_S^{}
        \left[
        \sum\limits_{i=1}^{3}
        (-1)^{i+1}
        \rho_i
        \sin{\alpha_i}
        \sin{\gamma_i}
        \right]^2,
    \end{aligned}
    \end{equation}
    where $S = \{(j_1,j_2,j_3) \; \vert \;  1 \leq j_1 < j_2 < j_3 \leq k \}$,
    \begin{equation*}
    \begin{aligned}
        &\alpha_i = \atantwo 
        \left(
        [\mathbf{u}_p \times \mathbf{u}_q]_z, \;
        \mathbf{u}_p \cdot \mathbf{u}_q
        \right),\\
        &{\gamma_i = \frac{\pi}{2} - \measuredangle 
        \left(
        \UnitZ \times \AlignedPosTarg_l, \mathbf{u}_l
        \right)}, 
    \end{aligned}
    \end{equation*}
    with ${l=j_i}$, $(p,q) \in \{j_1,j_2,j_3\} \setminus l$, $p < q$, and $[\cdot]_z$ extracts the $z$ element of the argument vector in $\mathbb{R}^3$.
    \item \label{case:2D_known_theta} 2D RTE with known $\theta$, $\StateVector \coloneqq [t^x, t^y]^\top$,
    \begin{equation}
    \begin{aligned}
        &\det(\FIM) = 
        \frac{1}{\sigma_r^2} 
        \sum\limits_{ 1 \leq i < j \leq k}^{}
        \sin^2{\alpha},
    \end{aligned}
    \end{equation}
    where $\alpha = \measuredangle (\mathbf{u}_i,\mathbf{u}_j)$.
\end{enumerate}
\end{prop}

\begin{proof}
    See Appendix \ref{appendix:sub_problems_derivation}.
\end{proof}

\subsection{Geometric interpretation of det(F)} \label{subsec:CRLB_geometric_interpretation}

\begin{figure}[t]
    \begin{subfigure}[t]{.495\linewidth}
    \centering
    \includegraphics[width=\linewidth]{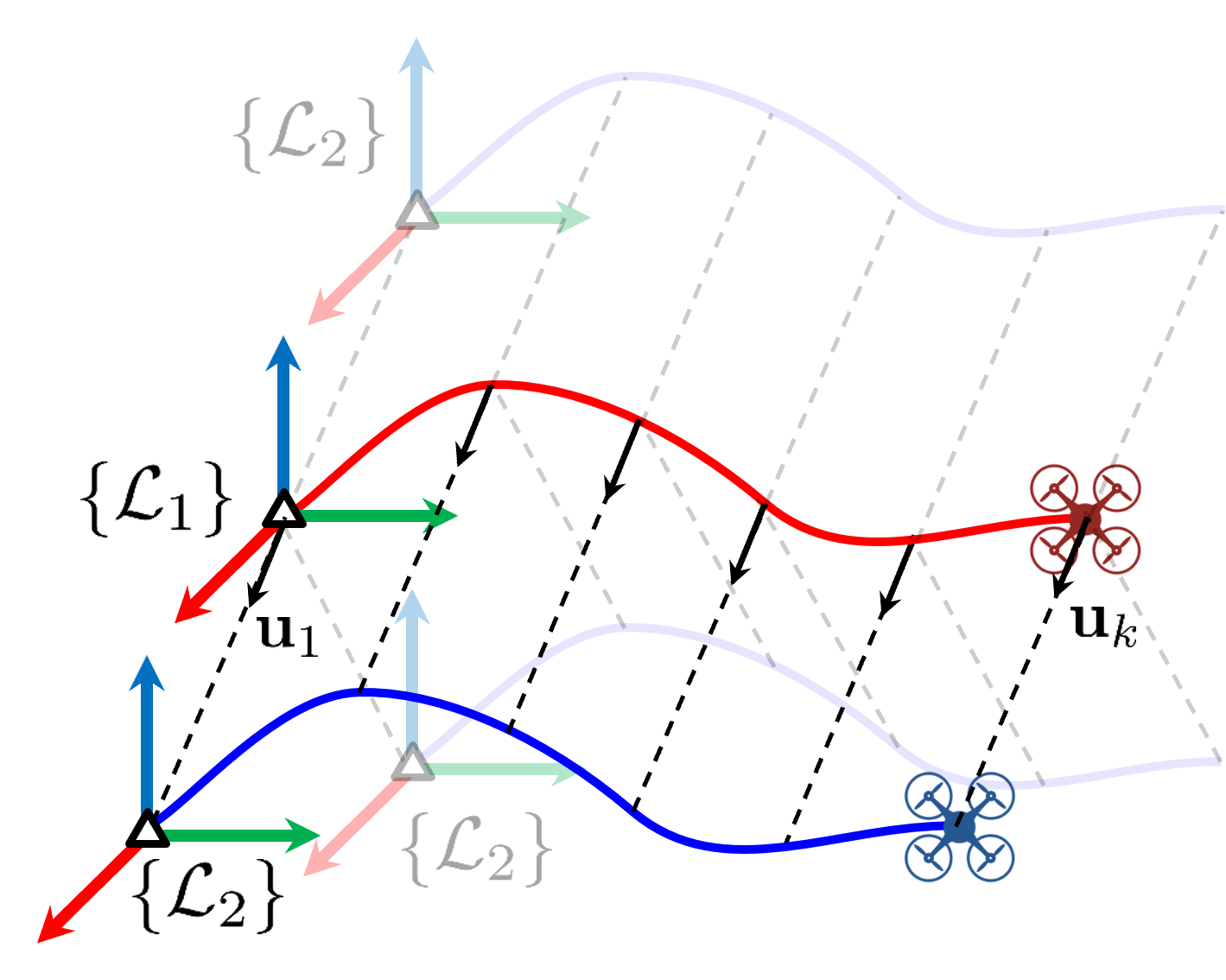}
    \caption{Robots move in parallel.}
    \label{fig:exp_singular_parallel}
    \end{subfigure}
    \hfill
    \begin{subfigure}[t]{.495\linewidth}
    \centering
    \includegraphics[width=\linewidth]{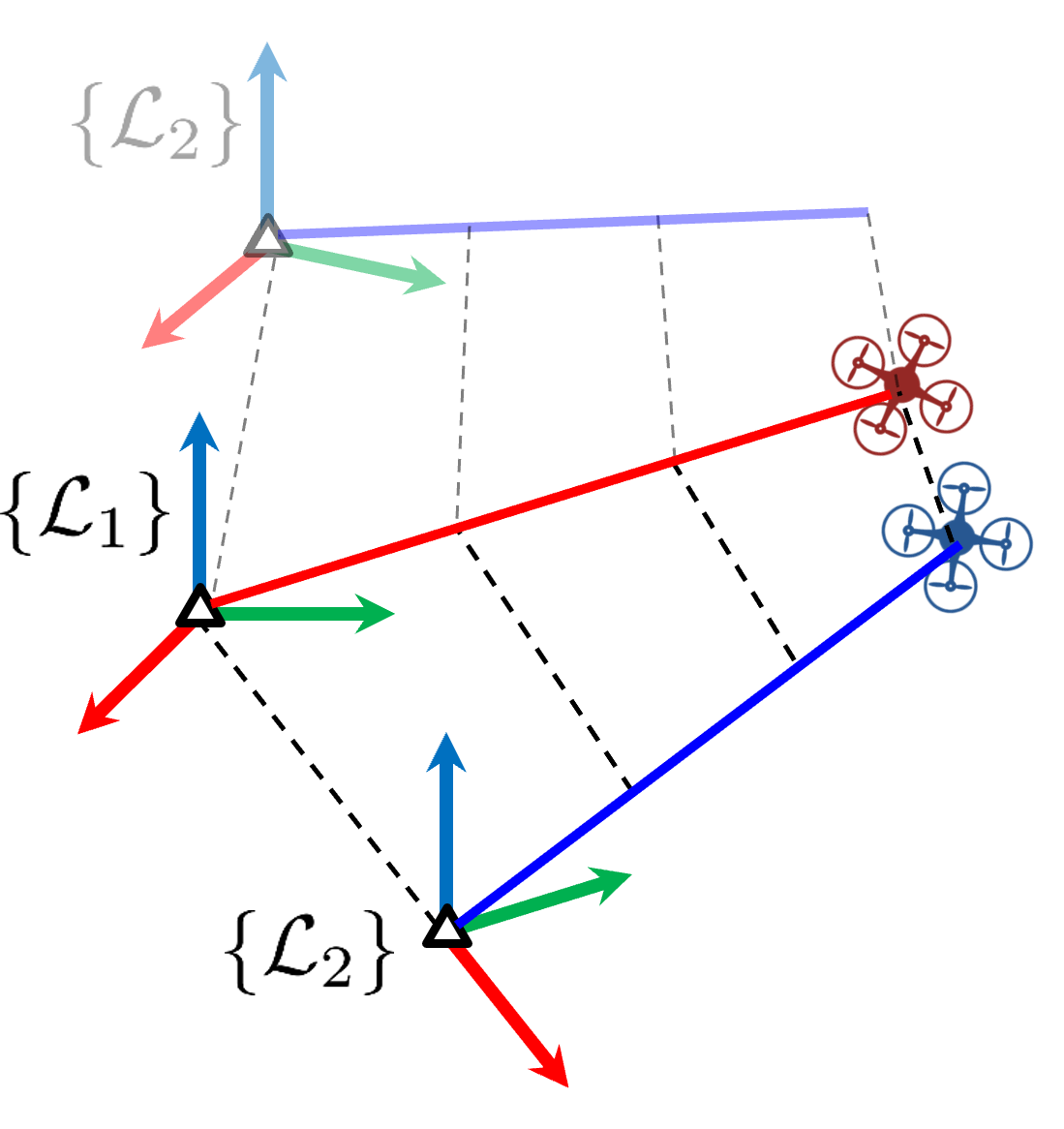}
    \caption{Robots move linearly on the same 2D plane.}
    \label{fig:exp_singular_planar}
    \end{subfigure}
    \begin{subfigure}[t]{0.495\linewidth}
    \centering
    \includegraphics[width=\linewidth]{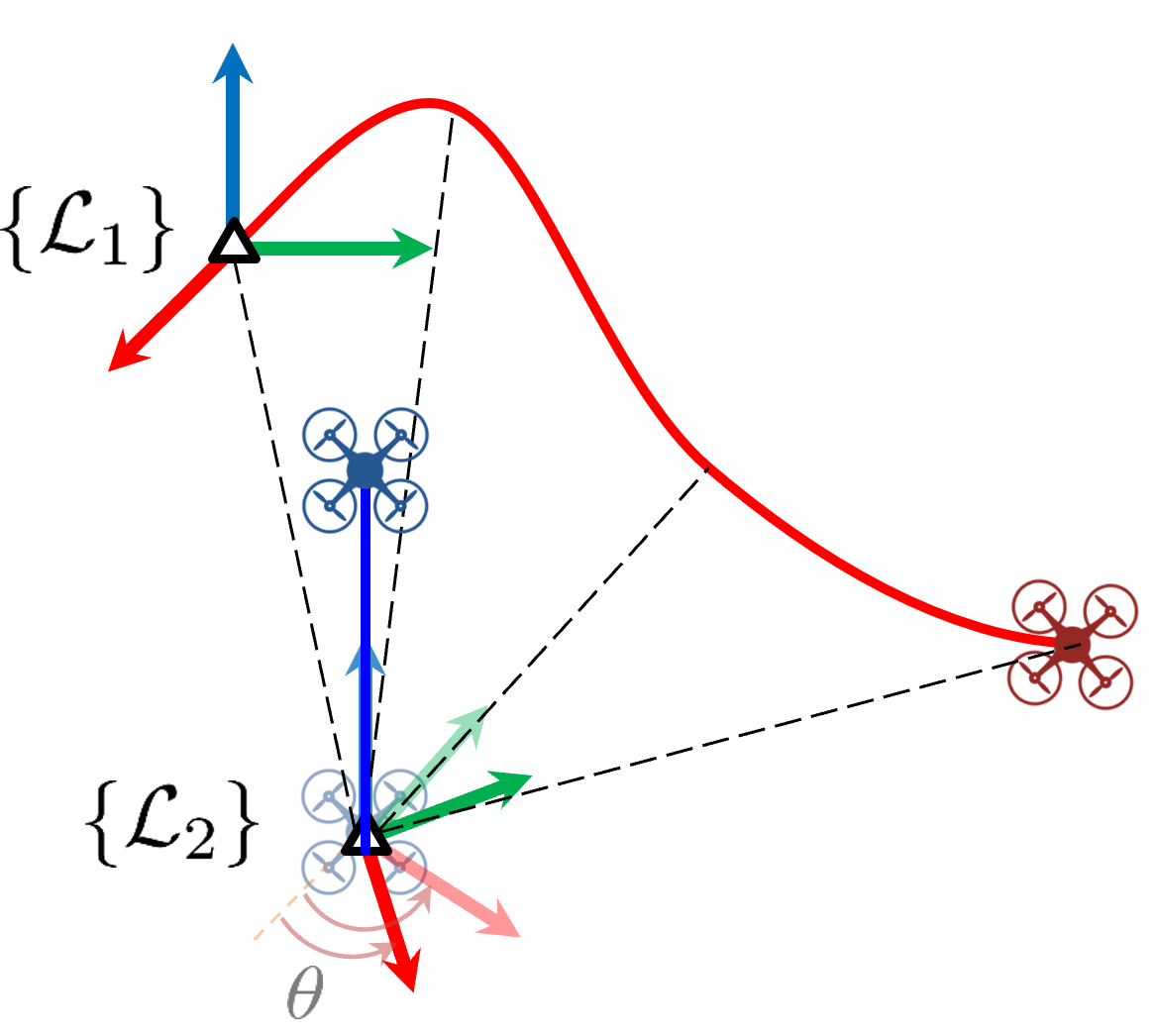}
    \caption{Target robot is stationary or moves only on the $z$ axis.}
    \label{fig:exp_singular_static_target}
    \end{subfigure}
    \hfill
    \begin{subfigure}[t]{.495\linewidth}
    \centering
    \includegraphics[width=\linewidth]{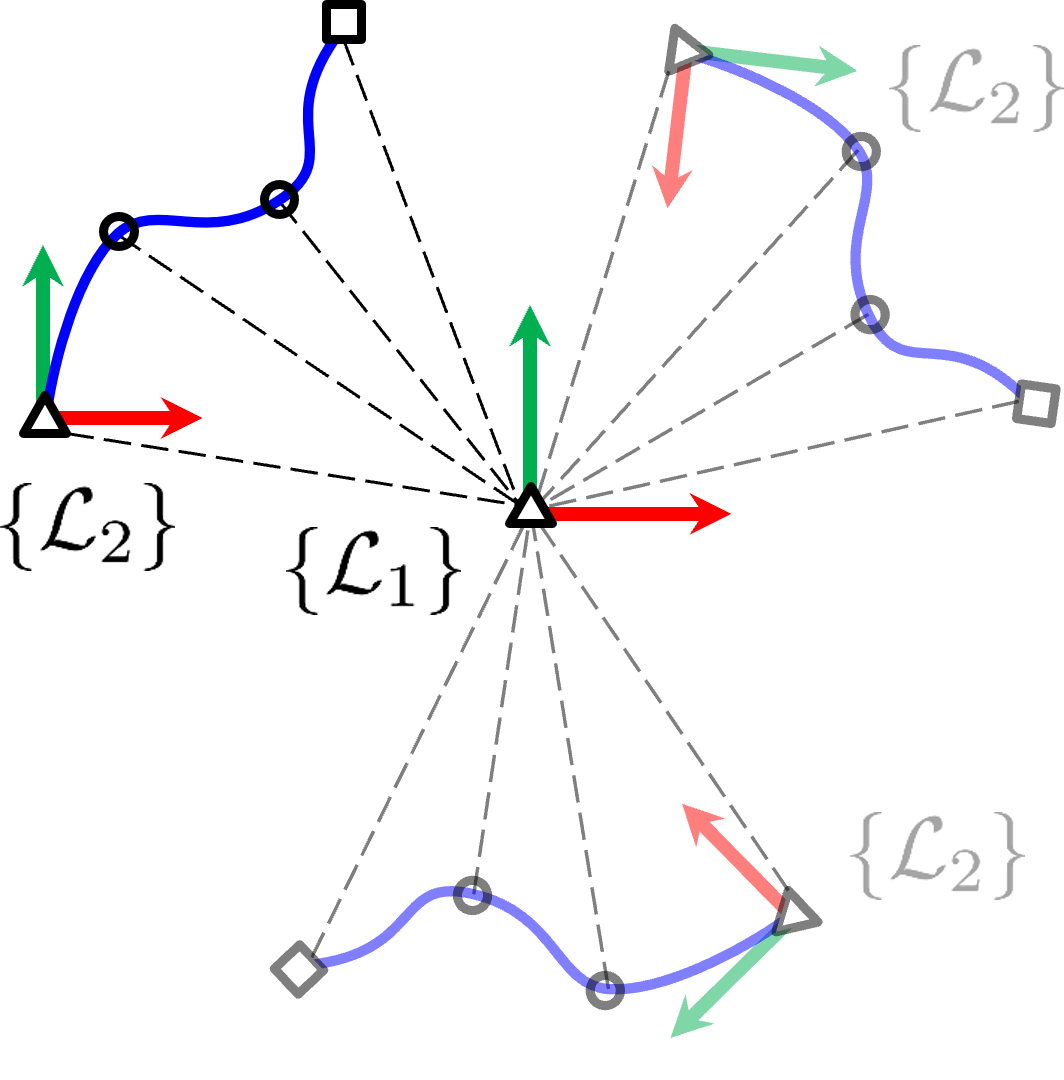}
    \caption{Host robot is stationary.}
    \label{fig:exp_singular_static_host}
    \end{subfigure}
    
\caption{Examples of singular configurations that lead to $\detF = 0$. The true and plausible solutions are depicted in solid and transparent colors, respectively.}
\label{fig:exp_degenerate_configs}
\end{figure}

We can interpret the components that build up $\detF$ in Eq. (\ref{eq:det_FIM_RTE}-\ref{eq:det_FIM_components}) as follows: $\mathbf{u}_m$ is the unit vector parallel to the relative position vector in the world frame $\{\mathcal{L}_1\}$ (Fig. \ref{fig:analytical_coord_systems}a). 
$\AlignedPosTarg_m$ is the local position vector of the target robot $\Robot_2$ as measured in the world frame. 
$T_i$ is the signed volume of the parallelepiped with three vectors $\mathbf{u}_l$, $\mathbf{u}_p$, $\mathbf{u}_q$ as edges (Fig. \ref{fig:viz_T_i}). 
$\Phi_i$ is the signed volume of the parallelepiped with three vectors $\AlignedPosTarg_m$, $\mathbf{u}_m$ and $\UnitZ$ as edges (Fig. \ref{fig:analytical_coord_systems}c). Essentially, for every set of four measurements at $j_1,\cdots,j_4$, $\detL$ is a combination of both $T_i$ and $\Phi_i$. Finally, $\detF$ is computed over all possible sets of four measurements $S$. The larger $\detF$ is, the smaller the uncertainty volume \cite{ly2017FIMtutorial}.

Intuitively, $T_i = \abs{\sin\alpha_i} \sin\beta_i$ represents the information gain regarding the translation parameters ($t^x, t^y, t^z$) and is directly correlated with the volume occupied by the relative position vectors (encoded by $\alpha_i$ and $\beta_i$) but not the absolute positions of the robots. This suggests that in order to improve the estimates of $\mathbf{t}$, one should focus on enlarging the angles between the measurements, which is exactly the strategy that was presented in \cite{guo2017reloc} for a static target. 
On the other hand, $\Phi_i = \rho_i \sin\gamma_i$ represents the information gain regarding the relative heading parameter ($\theta$) and is directly influenced by the horizontal displacement of the target robot in its local frame ($\rho_i$) and how perpendicular is the relative position vector ($\mathbf{u}_m$) to the vertical plane that contains $\AlignedPosTarg_m$ (which has the normal vector $\AlignedPosTarg_m \times \UnitZ$), measured by $\sin\gamma_i$. 

For a given configuration, if $T_i = 0$ or $\Phi_i = 0$ or $\textstyle \sum_{i=1}^{4} (-1)^{i} \; \Phi_i T_i {=} 0$ for the whole set $S$, then $\detF = 0$, in which case the associated parameters are at the critical points of $\mathbf{f}(\StateVector)$ and their uncertainty cannot be reduced regardless of the number of measurements. For example:
\begin{enumerate}
    \item If the robots move in parallel, so do the relative position vectors, i.e. $\mathbf{u}_l \: \| \: \mathbf{u}_p$ or $\alpha_i = 0 \; \forall i$, then $T_i = 0 \; \forall i$ and $\detF=0$. In this case, the relative translation $\mathbf{t}$ cannot be fully resolved (Fig. \ref{fig:exp_singular_parallel}).
    \item If the trajectories of the robots are linear and on the same 2D plane then their relative position vectors are also on the same 2D plane, which leads to $(\mathbf{u}_l \times \mathbf{u}_p) \perp \mathbf{u}_q$ or equivalently $\beta_i = 0 \; \forall i$, hence $T_i = 0 \; \forall i$ and $\detF=0$. In this case, the solution can be recovered up to a flip ambiguity (Fig. \ref{fig:exp_singular_planar}).
    \item If the target robot is stationary or only moves on the $z$ axis, i.e. $\LocPosTarg_m = \mathbf{0}$ or $\rho_i = 0$, then $\Phi_i = 0 \; \forall i$ and $\detF=0$. In this case, the relative heading $\theta$ cannot be fully resolved (Fig. \ref{fig:exp_singular_static_target}).
    \item If the host robot is stationary, then $\textstyle \sum_{i=1}^{4} (-1)^{i} \; \Phi_i T_i = 0$ and $\detF=0$. In this case, both $\mathbf{t}$ and $\theta$ cannot be fully resolved since the solution is invariant to rotation around the host robot (Fig. \ref{fig:exp_singular_static_host}).
\end{enumerate}

These examples with clear physical interpretations have also been stated separately in \cite{van2019board,cornejo2015distributed}. Additionally, we provide simulation results in Sect. \ref{subsec:UncertaintyEval} to demonstrate the above observations. 
\textcolor{black}{Besides identifying degenerate cases, $\detF$ can also be used to study the optimal configurations and online trajectory planning. Specifically, from a given number of measurements, by maximizing $\det(\mathbf{F})$ (or equivalently minimizing the uncertainty volume) subjected to limited sensing range and other motion constraints, we can find the optimal configurations. On the other hand, the online trajectory planning problem can use the D-optimal cost function ($J = -\ln\det(\mathbf{F})$) to plan the next best sensing positions given the current estimates. Although these studies are outside the scope of this paper, they are interesting subjects for future works.
}

\section{Proposed Approaches} \label{sec:main_approaches}

In this section, the proposed methods are presented in details. First, we briefly summarize the squared distances weighted least squares (SD-WLS) problem \cite{trawny2010relplanar}. Then, the SD-WLS problem is reformulated as a non-convex QCQP problem and further as an SDP relaxation problem.

\subsection{4-DoF squared distances weighted least squares} \label{subsec:main_approaches_SD_WLS}

The true inter-robot distance at time $t_k$ can be written as
\begin{equation}
    d_k = \norm{\mathbf{w}_k} = \sqrt{\mathbf{w}_k^\top \mathbf{w}_k},
\end{equation}
where $\mathbf{w}_k \coloneqq \mathbf{t} + \AlignedPosTarg_k - \LocPosHost_k$. The ranging error vector is
\begin{equation}
    \mathbf{e}_r \coloneqq \left[
        \norm{\mathbf{w}_1} - \tilde{d}_1,\;\;
        \norm{\mathbf{w}_2} - \tilde{d}_2,\;
        \cdots,\;
        \norm{\mathbf{w}_k} - \tilde{d}_k
    \right]^\top
\end{equation}
The squared distance measurement 
\begin{equation}
    \tilde{d}_k^2 = d_k^2 + 2 d_k \eta_k + \eta_k^2
\end{equation}
has the noise term $\nu_k \coloneqq 2 d_k \eta_k + \eta_k^2$, which is not zero-mean Gaussian. However, this non-Gaussian pdf can be approximated as one \cite{trawny2010relplanar}:
\begin{equation}
\begin{aligned}
    &\tilde{s}_k = d_k^2 + \eta'_k = \mathbf{w}_k^\top \mathbf{w}_k + \eta'_k,\\
    &\tilde{s}_k \simeq \tilde{d}_k^2 - \bar{\nu}_k, \;\; \bar{\nu}_k \coloneqq \mathbb{E}[\nu_k] = \CovMatR_{k,k},\\
    &\bm{\eta}' = [\eta'_1, \dots, \eta'_k] \sim \mathcal{N}(\bm{\eta}'; \mathbf{0}, \CovMatS).
\end{aligned}
\end{equation}
The elements of the covariance matrix $\CovMatS$ are computed as
\begin{equation}
\begin{aligned} \label{eq:CovSigmaS}
    \CovMatS_{i,i} &\coloneqq 
    \mathbb{E}[(\nu_i - \bar{\nu}_i)^2] = \CovMatR_{i,i} 
    \left(4 \tilde{d}_i^2 + 2 \CovMatR_{i,i}\right),\\
    \CovMatS_{i,j} &\coloneqq 
    \mathbb{E}[(\nu_i - \bar{\nu}_i)(\nu_j - \bar{\nu}_j)] 
    = \CovMatR_{i,j} 
    \left(4 \tilde{d}_i \tilde{d}_j  + 2\CovMatR_{i,j}\right).
\end{aligned}
\end{equation}

The 4-DoF SD-WLS problem is defined as
\begin{equation}\label{eq:prob_SD_WLS}
    \min\limits_{\StateVector} \frac{1}{2}
        \mathbf{e}_s^\top
        \CovMatS^{-1}
        \mathbf{e}_s
\end{equation}
where $\mathbf{e}_s$ is the vector of the squared distance errors
\begin{equation} \label{eq:e_s}
    \mathbf{e}_s \coloneqq [
    \mathbf{w}_1^\top \mathbf{w}_1 - \tilde{s}_1, \;\;
    \mathbf{w}_2^\top \mathbf{w}_2 - \tilde{s}_2, \dots , \;
    \mathbf{w}_k^\top \mathbf{w}_k - \tilde{s}_k]^\top.
\end{equation}
The original SD-WLS \cite{trawny2010relplanar} was established for the 2D case. Here, we change the parameters to that of the 4-DoF case.

\subsection{Non-convex QCQP} \label{subsec:main_approaches_QCQP}

Expanding and simplifying $\mathbf{w}_k^\top \mathbf{w}_k$ lead to:
\begin{equation}\label{eq:squared_dist_full_form}
\begin{aligned}
    &\mathbf{w}_k^\top \mathbf{w}_k = (\mathbf{t} + \mathbf{C} \LocPosTarg_k - \LocPosHost_k)^\top (\mathbf{t} + \mathbf{C} \LocPosTarg_k - \LocPosHost_k)\\
    &= \mathbf{t}^\top \mathbf{t} + \LocPosHost_k^\top \LocPosHost_k + \LocPosTarg_k^\top \LocPosTarg_k \\
    &+ 2 (\mathbf{t} - \LocPosHost_k)^\top \mathbf{C} \LocPosTarg_k - 2 \LocPosHost_k^\top \mathbf{t}.
\end{aligned}
\end{equation}

Let $\mathbf{x} = [x_1,x_2,\dots,x_9]^\top$ be the state vector:
\begin{equation} \label{eq:x_state_vector}
\begin{aligned}
    &\mathbf{x} \coloneqq [
    t^x, \;\;
    t^y, \;\;
    t^z, \;\; 
    \cos{\theta},\;\; 
    \sin{\theta},\;\; 
    t^x \cos{\theta} + t^y \sin{\theta}, \\
    &t^y \cos{\theta} - t^x \sin{\theta}, \;\;
    (t^x)^2 + (t^y)^2 + (t^z)^2, \;\; 
    1]^\top \in \R^{9 \times 1},
\end{aligned}
\end{equation}
each row of $\mathbf{e}_s$ can be rearranged as $\mathbf{w}_i^\top \mathbf{w}_i - \tilde{s}_i = \mathbf{A}_i \mathbf{x}$, where
\textcolor{black}{
\begin{equation}
\begin{aligned} \label{eq:Ai_row}
    \mathbf{A}_i &\coloneqq [
    {-}2\varphi^x_i, \;
    {-}2\varphi^y_i, \;
    2(o^z_i {-} \varphi^z_i), \;
    {-}2(\varphi^x_i o^x_i {+} \varphi^y_i  o^y_i), \\
    &2(\varphi^x_i  o^y_i {-} o^x_i \varphi^y_i), \;\;
    2  o^x_i, \;\;
    2  o^y_i, \;\;
    1, \;\;
    \varepsilon_i] \in \R^{1 \times 9},\\
    \varepsilon_i &\coloneqq 
    \LocPosHost_i^\top \LocPosHost_i + \LocPosTarg_i^\top \LocPosTarg_i - 2\varphi^z_i o^z_i - \tilde{s}_i,\\
\end{aligned}
\end{equation}
}
with $i = 1,\dots,k$.
$\mathbf{e}_s$ in Eq. (\ref{eq:e_s}) can then be simplified as
\begin{equation}
    \mathbf{e}_s = 
        \begin{bmatrix}
            \mathbf{A}_1\\
            \vdots\\
            \mathbf{A}_k
        \end{bmatrix}
        \mathbf{x}
    = \mathbf{B} \mathbf{x},
\end{equation}
which allows the SD-WLS cost function (\ref{eq:prob_SD_WLS}) to be reformulated into a quadratic form
\begin{equation}
\begin{aligned}
    \frac{1}{2} \mathbf{e}_s^\top \CovMatS^{-1} \mathbf{e}_s = 
    \frac{1}{2} \mathbf{x}^\top 
    \overbrace{
    \mathbf{B}^\top \CovMatS^{-1} \mathbf{B}
    }^{\mathbf{P}_0}
    \mathbf{x} = \frac{1}{2} \mathbf{x}^\top \mathbf{P}_0 \mathbf{x}.
\end{aligned}
\end{equation}

Finally, the 4-DoF SD-WLS problem (\ref{eq:prob_SD_WLS}) is reformulated as a non-convex QCQP problem
\begin{equation}\label{eq:prob_QCQP}
\begin{aligned}
    \min\limits_{\mathbf{x}} \;\; & \mathbf{x}^\top \mathbf{P}_0 \mathbf{x}\\
    \textrm{s.t.} \;\; 
    &\mathbf{x}^\top \mathbf{P}_i \mathbf{x} = r_i,\;\;i=1,\dots,5.\\
\end{aligned}
\end{equation}

The constraints are drawn from the relations between the parameters of $\mathbf{x}$ in Eq. (\ref{eq:x_state_vector}) as follows:
\begin{equation}
\begin{aligned}
    &\cos^2\theta + \sin^2\theta = 1 
        \Leftrightarrow
    x_4^2 + x_5^2 = 1
    \Rightarrow \mathbf{x}^\top \mathbf{P}_1 \mathbf{x} = r_1,\\
    &t^x \cos{\theta} + t^y \sin{\theta} = x_6 x_9 
        \Leftrightarrow
    x_1 x_4 {+} x_2 x_5 {-} x_6 x_9 {=} 0 \\
    &\Rightarrow \mathbf{x}^\top \mathbf{P}_2 \mathbf{x} = r_2,\\
    &t^y \cos{\theta} - t^x \sin{\theta} = x_7 x_9 
        \Leftrightarrow
    x_2 x_4 - x_1 x_5 - x_7 x_9 {=} 0 \\
    &\Rightarrow \mathbf{x}^\top \mathbf{P}_3 \mathbf{x} = r_3,\\
    &(t^x)^2 + (t^y)^2 + (t^z)^2 {=} x_8 x_9 
        \Leftrightarrow
    x_1^2 + x_2^2 + x_3^2 - x_8 x_9 {=} 0 \\
    &\Rightarrow \mathbf{x}^\top \mathbf{P}_4 \mathbf{x} = r_4,\\
    &(t^x)^2 + (t^y)^2 + (t^z)^2 = d_0^2 
        \Leftrightarrow
    x_1^2 + x_2^2 + x_3^2 = \tilde{d}_0^2 \\
    & \Rightarrow \mathbf{x}^\top \mathbf{P}_5 \mathbf{x} = r_5,
\end{aligned}
\end{equation}
where
\begin{equation}
\begin{aligned} \label{eq:QCQP_constraints}
    &\mathbf{P}_1 = \operatorname{sparse}([4,5],[4,5],[1,1],9,9),
    r_1 = 1,\\ 
    &\mathbf{P}_2 = \operatorname{sparse}([1,2,9],[4,5,6],[1,1,-1],9,9), 
    r_2 = 0,\\  
    &\mathbf{P}_3 = \operatorname{sparse}([2,1,9],[4,5,7],[1,-1,-1],9,9), 
    r_3 = 0,\\  
    &\mathbf{P}_4 = \operatorname{sparse}([1,2,3,8],[1,2,3,9],[1,1,1,-1],9,9), 
    r_4 = 0,\\  
    &\mathbf{P}_5 = \operatorname{sparse}([1,2,3],[1,2,3],[1,1,1],9,9),
    r_5 = \tilde{d}_0^2. 
\end{aligned}
\end{equation}
Here, we use the MATLAB operator $\operatorname{sparse}(\mathbf{a},\mathbf{b},\mathbf{c},m,n)$ which generates an $m \times n$ sparse matrix $\mathbf{P}$ from the vectors $\mathbf{a}$, $\mathbf{b}$, and $\mathbf{c}$, such that $\mathbf{P}_{a_k,b_k} = c_k$.

\begin{rem}\label{rem:SD_WLS_d0}
While it is possible to reduce the number of parameters by exploiting $\tilde{d}_0$\cite{li2020relSDP}, we found that in real-life scenarios $\tilde{d}_0$ might not exist (e.g., if one robot starts moving when the other robot is not ready for operation, or if the starting points are not within line-of-sight). Hence, $\tilde{d}_0$ is used only for constraint $i{=}5$, which is removed if $\tilde{d}_0$ is not available.
\end{rem}

\subsection{SDP Relaxation} \label{subsec:main_approaches_SDP}

\begin{algorithm}
\color{black}
\caption{\color{black}SDP relaxation algorithm for $N=2$ robots}\label{alg:sdp}
\begin{algorithmic}
    \Require Inter-robot ranging measurements $\{\tilde{d}_i\}_{i=1}^k$, self-odometry measurements $\{\prescript{\mathcal{L}_1}{a_1}{\tilde{\mathbf{{p}}}_i}$, $\prescript{\mathcal{L}_2}{a_2}{\tilde{\mathbf{{p}}}_i}\}_{i=1}^k$.
    \Ensure Relative frame transformation $\hat{\StateVector}$
    \State Setup $\{\mathbf{P}_i, r_i\}_{i=1}^5$ according to Eq. (\ref{eq:QCQP_constraints})
    \State $i \gets 1$
    \While{$i \leq k$}
        \State Calculate the $i$-th row of $\CovMatS$ according to Eq. (\ref{eq:CovSigmaS})
        \State Calculate the $i$-th row of $\mathbf{B}$ according to Eq. (\ref{eq:Ai_row})
        \State $i \gets i + 1$
    \EndWhile
    \State $\mathbf{P}_0 \gets \mathbf{B}^\top \CovMatS^{-1} \mathbf{B}$
    \State $\overset{*}{\mathbf{X}} \gets$ solve the SDP problem (\ref{eq:prob_SDP}) given $\mathbf{P}_0$, $\{\mathbf{P}_i, r_i\}_{i=1}^5$
    \State $[\mathbf{U}, \mathbf{S}, \mathbf{V}] \gets \textrm{svdsketch}(\overset{*}{\mathbf{X}})$
    \State \textbf{if} $\textrm{rank}(\overset{*}{\mathbf{X}}) == 1$ \textbf{then} $\overset{*}{\mathbf{x}} \gets \mathbf{U}$ \textbf{else} $\overset{*}{\mathbf{x}} \gets \sqrt{S_{1,1}} \mathbf{U}_{1:9,1}$
    \State \textbf{if} $\overset{*}{x}_9 < 0$ \textbf{then} $\overset{*}{\mathbf{x}} = -\overset{*}{\mathbf{x}}$
    \State $\hat{\mathbf{t}} \gets \overset{*}{\mathbf{x}}_{1:3}$, $\hat{\theta} \gets \atantwo(\overset{*}{x}_5, \overset{*}{x}_4)$
    \State $\hat{\StateVector} \gets [\hat{\mathbf{t}}^\top, \hat{\theta}]^\top$
\end{algorithmic}
\end{algorithm}

\subsubsection{Problem reformulation}

We follow the steps outlined in \cite{luo2010sdp} to obtain the SDP relaxation formulation of the QCQP problem (\ref{eq:prob_QCQP}). First, observe that 
\begin{equation}
    \mathbf{x}^\top \mathbf{P}_i \mathbf{x} = \operatorname{Tr}( \mathbf{x}^\top \mathbf{P}_i \mathbf{x}) = \operatorname{Tr}( \mathbf{P}_i \mathbf{x} \mathbf{x}^\top).
\end{equation}
By using a new variable $\mathbf{X} = \mathbf{x} \mathbf{x}^\top$, an equivalent formulation of (\ref{eq:prob_QCQP}) can be written as:
\begin{equation}\label{eq:prob_equivalent_QCQP}
\begin{aligned}
    \min\limits_{\mathbf{X}} \;\; &\operatorname{Tr}( \mathbf{P}_0 \mathbf{X})\\
    \textrm{s.t.} \;\;
    &\operatorname{Tr}( \mathbf{P}_i \mathbf{X} ) = r_i,\;\;i=1,\dots,5\\
    &\; \mathbf{X} \succeq 0,\;\operatorname{rank} (\mathbf{X}) = 1,
\end{aligned}
\end{equation}
where $\mathbf{X} \succeq 0$ indicates that $\mathbf{X}$ is positive semidefinite. By dropping the nonconvex rank constraint, we obtain the SDP relaxation problem as
\begin{equation}\label{eq:prob_SDP}
\begin{aligned}
    \min\limits_{\mathbf{X}} \;\; &\operatorname{Tr}( \mathbf{P}_0 \mathbf{X})\\
    \textrm{s.t.} \;\;
    &\operatorname{Tr}( \mathbf{P}_i \mathbf{X} ) = r_i,\;\;i=1,\dots,5\\
    &\; \mathbf{X} \succeq 0.
\end{aligned}
\end{equation}

The advantage of the SDP problem (\ref{eq:prob_SDP}) over the original NP-hard problems (\ref{eq:prob_QCQP}) and (\ref{eq:prob_equivalent_QCQP}) is that it can be solved in polynomial time. 
In the noiseless or low noise cases, the SDP relaxation can be expected to be tight, i.e. solving either the relaxed and original problems is equivalent \cite{luo2010sdp}. A problem specific explanation which is also applicable to our case can be found in Lemma 1 in \cite{li2020relSDP}.

\subsubsection{Recover the original solution}

Let $\overset{*}{\mathbf{x}}$ and $\overset{*}{\mathbf{X}}$ be the solution of (\ref{eq:prob_QCQP}) and (\ref{eq:prob_SDP}), respectively. If $\textrm{rank}(\overset{*}{\mathbf{X}}) = 1$, then the low-rank decomposition $\overset{*}{\mathbf{X}} = \overset{*}{\mathbf{x}} (\overset{*}{\mathbf{x}})^\top$ will provide the feasible and optimal solution $\overset{*}{\mathbf{x}}$. If $\textrm{rank}(\overset{*}{\mathbf{X}}) > 1$, the rank-one decomposition process in \cite{luo2010sdp} can be used to extract $\overset{*}{\mathbf{x}}$. Using SVD to decompose $\overset{*}{\mathbf{X}}$ as
\begin{equation}
    \overset{*}{\mathbf{X}} = \sum\limits_{i=1}^{n} \lambda_i \mathbf{q}_i \mathbf{q}_i^\top,\; \lambda_1 \geq \lambda_2 \geq \dots \geq \lambda_n > 0,
\end{equation}
where $\lambda_i$ and $\mathbf{q}_i$ are the eigenvalues and respective eigenvectors, the best rank-one approximation is
\begin{equation}
    \overset{*}{\mathbf{X}}_1 = \lambda_1  \mathbf{q}_1  \mathbf{q}_1^\top,
\end{equation}
Lastly, since we expect $x_9$ to be positive from the way $\mathbf{x}$ is constructed (Eq. (\ref{eq:x_state_vector})), the sign of the final solution is flipped if ${\overset{*}{x}_9}$ is negative. \textcolor{black}{Our method is summarized in Algorithm \ref{alg:sdp}.}

\subsection{Practical issues consideration}
\subsubsection{UWB outliers rejection} \label{subsubsec:good_practice_uwb_outlier}


We employ the same UWB outlier rejection scheme as our previous work \cite{Thien2021multiviro}. The variance of the UWB data in a sliding window consisting of $K$ latest samples is computed as
\begin{equation}
    \hat{\sigma}_k^2 = \frac{1}{K}\sum\limits_{i=k-K+1}^{k}(\tilde{d}_i - \bar{d}_k)^2,
\end{equation}
where $\bar{d}_k$ is the mean value of $K$ samples. When a new data $\tilde{d}_{k+1}$ is received, if the new $\hat{\sigma}_{k+1}$ increases over a pre-defined threshold, the data is discarded. In our real-life experiments, we set $K=20$ and the threshold as $0.005$.



\subsubsection{Singular configuration detection}\label{subsubsec:UncertaintyEst}
Singular configurations refer to the cases where multiple solutions exist for at least one of the parameters given the available measurements \cite{trawny2010rel3Dtransform}. In the physical sense, this happens if the target robot's trajectory can be flipped, rotated or a combination of both in the world frame and the distance measurements will be the same \cite{cornejo2015distributed}. This observation corroborates our interpretation of the intuitive unobservable cases as discussed previously in Sect. \ref{subsec:CRLB_geometric_interpretation}. In previous works listed in Table \ref{table:lit_review}, no detection method for singular configurations nor measures of the uncertainty of the estimates were provided. 

In the noiseless case, if one parameter is unobservable in a given configuration, the associated row and column in the FIM will become zeros (no information gain from the available measurements). As such, the singular configurations will manifest in the form of the FIM losing rank in noiseless cases, or the condition number of the FIM approaches infinity in noisy cases. Our detection scheme for singular configuration works as follows. Firstly, we evaluate the FIM using the analytical formula in Eq. (\ref{eq:FIM_simplified}) at the latest estimates $\hat{\StateVector}_k$. Then, we compute the condition number of the estimated FIM $\condF$. A configuration is deemed singular if $\kappa(\hat{\FIM}_k)$ is larger than a threshold, which can be determined empirically.

In practice, before the first optimization we check that the sample variance based on recent positions of the robots is higher than a threshold (empirically set as $0.05\si{m}$ for the last $100$ positions in our experiments), otherwise the optimization process is skipped. This simple check ensures that 1) both robots are moving, and 2) there are motions on all axes, which is sufficient to avoid simple singular cases such as one robot is static or both robots are moving in parallel. Additionally, it was observed that if the poses only change marginally in $x$ and $y$ directions, such as when the quadrotors are hovering, the estimates do not improve regardless of how many new measurements are added. As such, we keep performing this check during the mission and skip any unnecessary updates.

\begin{rem}
In the noiseless case, the singular configurations will also manifest in the form of matrix $\mathbf{P}_0$ in Eq. (\ref{eq:prob_QCQP}) losing rank, as the optimal solution of the QCQP problem is not unique. As a result, the singular cases can be detected by checking whether $\mathbf{P}_0$ is rank-deficient, which can avoid computing the FIM entirely. However, in our real-life experiments the noise will generally make $\mathbf{P}_0$ full rank even when the robots just move slightly. Hence, this method was not used.
\end{rem}

\subsubsection{Uncertainty of the estimates}
Let $\hat{\textrm{CRLB}}$ be the CRLB computed with the estimated $\hat{\StateVector}$ instead of the true $\StateVector$. Following \cite{ly2017FIMtutorial}, the 95\%-confidence interval for each parameter of $\StateVector$ is
\begin{equation}
    CI_i = \left[ 
        \hat{\Theta}_i - 1.96 \sqrt{\hat{\textrm{CRLB}}_{i,i}} \;, \;
        \hat{\Theta}_i + 1.96 \sqrt{\hat{\textrm{CRLB}}_{i,i}}
    \right]
\end{equation}
for $i \in \{1,\cdots,4\}$. Additionally, $\hat{\textrm{CRLB}}$, or generally the inverse of the observed FIM, can be used as the error covariance matrix of the Maximum Likelihood Estimator \cite{ponda2009trajopt}.

In essence, at time $t_k$, our framework reports the current estimates $\hat{\StateVector}$ and the corresponding uncertainty metrics, including the condition number $\kappa(\hat{\FIM}_k)$ and the standard error $\sigma_{\hat{\Theta}_i} \coloneqq \sqrt{\hat{\textrm{CRLB}}_{i,i}}$ for each parameter ($i \in \{1,\cdots,4\}$). If $\kappa(\hat{\FIM}_k)$ and $\sigma_{\hat{\Theta}_i}$ are sufficiently large, the trajectory configuration can be recognized as singular, with $\Theta_i$ being the unobservable parameter.

\subsection{Extension to multi-robot scenarios} \label{subsec:general_multi_robot}

{\color{black}
While we only focus on two-robot scenarios in this work, it is important to discuss the scalability of our method in general multi-robot cases. 
A simple approach of copying the same estimator to each pair of robots in the sensing graph is a valid solution and has been demonstrated in \cite{cornejo2015distributed,cossette2021relative}. However, in such scheme the computation burden on each robot will scale with the number of connected neighbors. Furthermore, there are no embedded mechanisms to ensure the consistency of the estimates across the network, combine the estimates to improve the global results or let the system run in an asynchronous and decentralized manner.
Hence, using a distributed optimization method designed for multi-robot systems \cite{halsted2021survey} would be a better choice.
In particular, the asynchronous-parallel Alternating Direct Method of Multipliers method introduced as part of the AROCK framework \cite{peng2016arock} is a fitting option for our problem. How to reformulate our QCQP problem (\ref{eq:prob_QCQP}) into AROCK and leverage the structure of our problem to improve the performance and reduce the computation demand is a worthy topic for future works.

With regard to communication, in \cite{ziegler2021distributed} it has been shown that the average traffic between two agents in an odometry and range-based relative formation system would be less than 2KB/s (using ROS). This would take up much less bandwidth than SOTA vision-based systems \cite{tian2021kimeramulti}, which can require MB/s per robot for keyframe data alone. As such, communication is not a critical issue with our method. However, the communication rate can be reduced even more by leveraging our findings.
From the analysis in Sect. \ref{subsec:CRLB_geometric_interpretation}, we have shown that there are motions during which the estimation results for some of the parameters would not improve (for quadrotors: hovering, yawing while hovering, ascending/ descending, etc.). As such, when the robot is moving in such manners it can stop sending data and the performance would not be affected. Hence, with our method each robot can self-access and decide whether broadcasting its local data is necessary.
}

\begin{figure*}[t]
\centering
	\includegraphics[width=\linewidth]{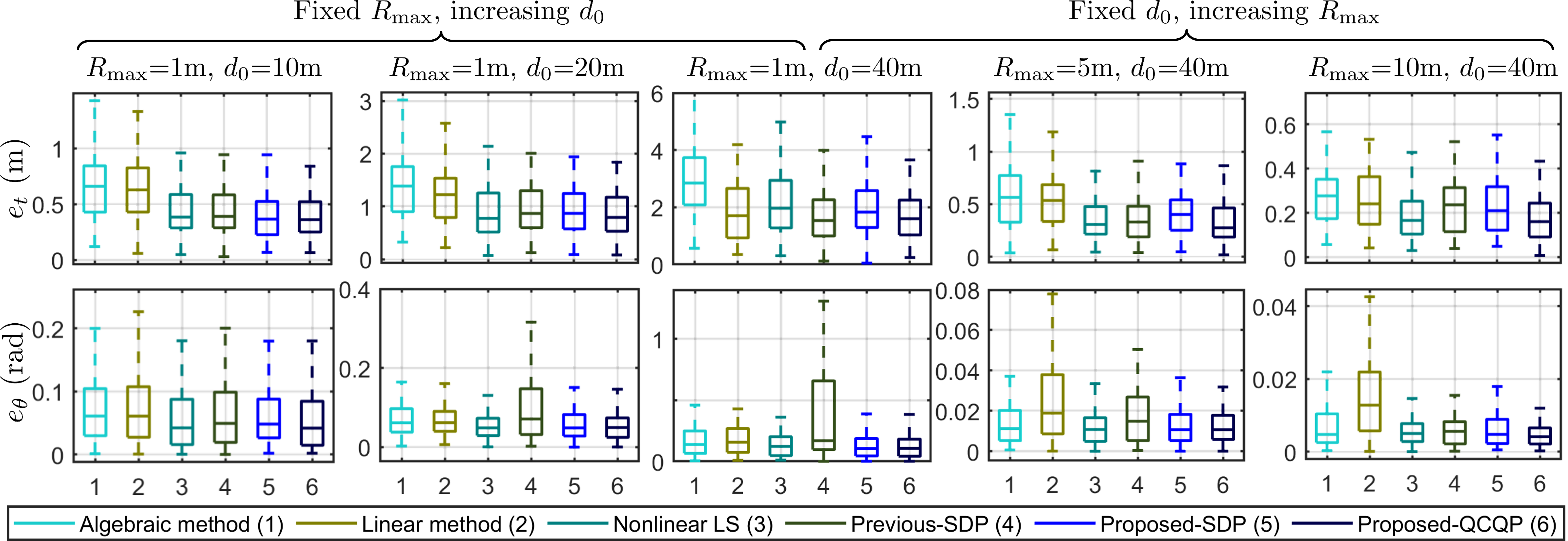}
    \caption{\textcolor{black}{Estimation errors with varying $R_{\max}$ and $d_0$. Top: translation error $e_t$. Bottom: heading error $e_{\theta}$. All simulations are run with $\sigma_o = 0.001\si{m}$ and $\sigma_r = 0.1\si{m}$. As $d_0$ increases or $R_{\max}$ decreases, the scale of the errors increases noticeably.}}
    \label{fig:compare_all_d0_and_Rmax}
\end{figure*}

\begin{figure*}[t]
\centering
	\includegraphics[width=\linewidth]{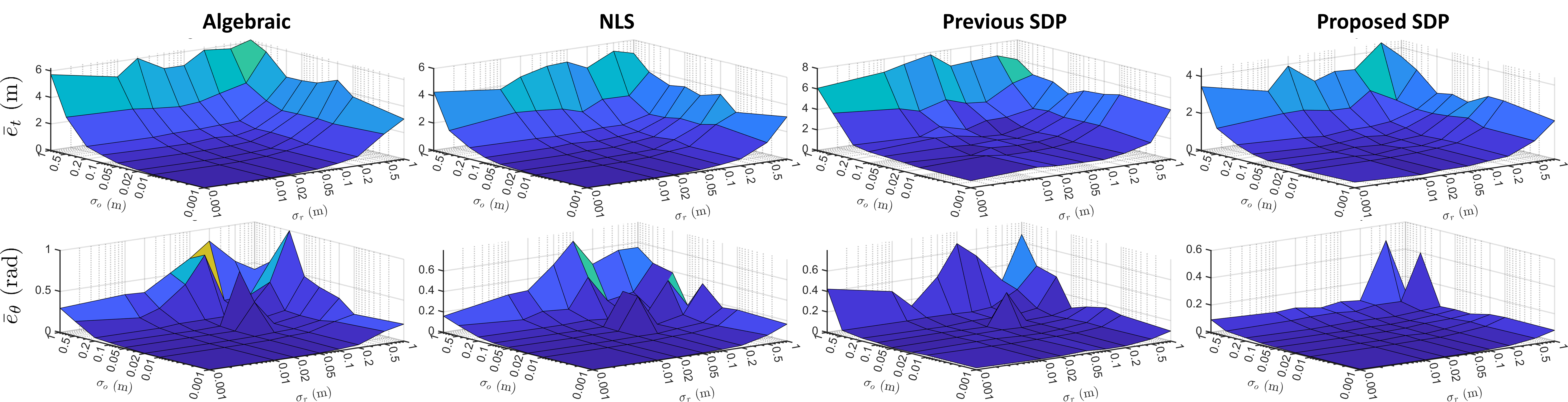}
    \caption{\textcolor{black}{Average translation $\bar{e}_t (\si{m})$ and heading $\bar{e}_{\theta} (\si{rad})$ errors with varying UWB noise ($\sigma_r$) and odometry noise ($\sigma_o$). All simulations are run with $d_0 = 50\si{m}$ and $R_{\max} = 10\si{m}$.}}
    \label{fig:compare_all_sigmas_vs_both_errors}
\end{figure*}

\begin{figure}[t]
\centering
	\includegraphics[width=\linewidth]{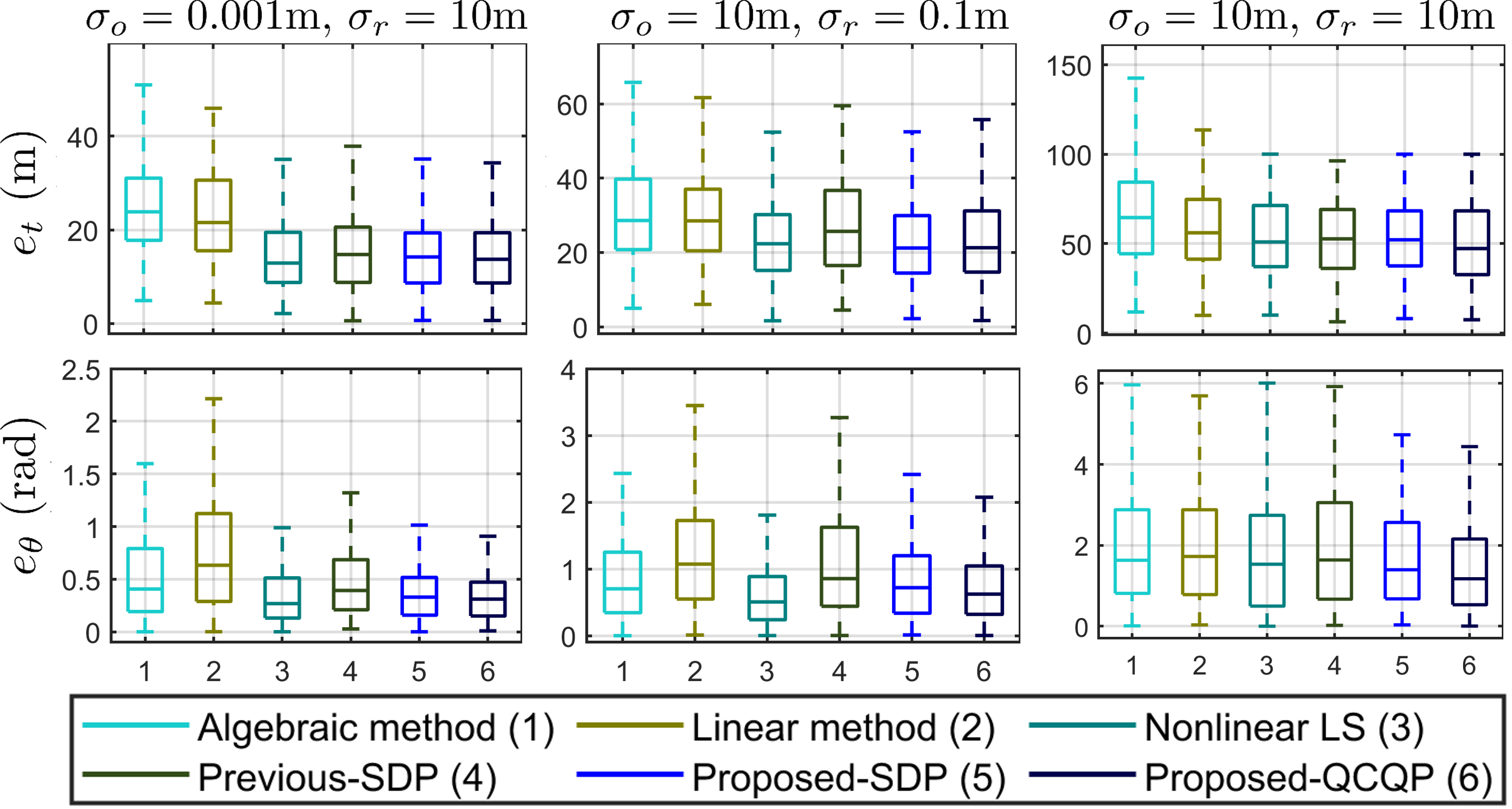}
    \caption{\textcolor{black}{Estimation errors under extreme noise levels. All simulations are run with $d_0 = 50\si{m}$ and $R_{\max} = 10\si{m}$.}}
    \label{fig:compare_all_vs_sigr_sigo}
\end{figure}

\section{Experimental Results} \label{sec:exp}

Video of the simulations and experiments can be found at \url{https://youtu.be/A_v-6fBEU_U}.

\subsection{Implementation details and performance metrics}

Among the 3D methods listed in Table \ref{table:lit_review}, we include the algebraic \cite{trawny2010rel3Dtransform}, linear \cite{molina2019unique}, nonlinear least squares (NLS) \cite{ziegler2021distributed} and previous SDP \cite{jiang2020rel3D} methods for comparison. None of these methods provide open-source code and in our experience, the implementation details can affect the performance noticeably. As such, we strive for fair comparison as follows:
\begin{itemize}
    \item For the methods in \cite{trawny2010rel3Dtransform,jiang2020rel3D} which are designed for full 6-DoF, we reduce the estimation problem to the same 4-DoF as ours by setting the extra parameters as constants and removing/adjusting the related constraints.
    \item For simulation, all methods are implemented in MATLAB with SeDuMi\footnote{\url{https://github.com/sqlp/sedumi}} solver for SDP methods. For real-life experiments, the methods are implemented in C++ with the solvers: CSDP\footnote{\url{https://github.com/coin-or/Csdp}} for SDP methods, Gurobi\footnote{\url{https://www.gurobi.com/}} for the QCQP method, Eigen\footnote{\url{https://eigen.tuxfamily.org/}} for algebraic and linear methods, Ceres\footnote{\url{http://ceres-solver.org/}} for the NLS method.
    \item The timing comparison takes into account only the solver time, i.e. the time it took to find the optimal solution, while excludes other data processing and preparation steps (construction and inversion of matrices, etc.).
    \item \textcolor{black}{Since the performance of NLS relies strongly on the initial guess, we tested the NLS method with a good starting point as follows. The guesses for $\mathbf{t}$ and $\theta$ are generated by adding Gaussian noises with $\sigma_t^0 = 0.5\si{m}$ and $\sigma_{\theta}^0 = 10\si{deg}$ to the ground truth values.}
    \item All methods are presented with the same processed input data (degenerate configurations checked, outlier rejection scheme applied, spatial-temporal offsets compensated).
\end{itemize}

We evaluate the translation and heading parameters separately, with the estimated translation and heading error denoted as $e_t = \norm{\hat{\mathbf{t}} - \mathbf{t}}$ and $e_{\theta} = \abs{\hat{\theta} - \theta}$, respectively.
The average errors over all runs $\bar{e}_t$ and $\bar{e}_{\theta}$, the root mean square error $\textrm{RMSE}_t$ and $\textrm{RMSE}_{\theta}$, the mean square errors (MSE) against the CRLBs, and the solver time are the subjects of comparison.

\subsection{Simulation}

The simulations are designed to evaluate different factors in isolation. We focus on two essential factors that can affect the performance: the trajectories configuration and the signal noises.
As previously mentioned, $d_0 = \norm{\mathbf{t}}$ is the true initial relative distance between the robots. 
Let $R_{\max}$ be the maximum moving radius of the robots from their initial positions, i.e. all possible trajectories are confined in a sphere centered at the local origin with a radius $R_{\max}$ (Fig. \ref{fig:sys_overview}a). The standard deviation of UWB and odometry data are denoted as $\sigma_r$ and $\sigma_o$, respectively.

In previous works, the simulations are done for a specific scenario: the trajectories' shape and size are fixed, and often only one value of $\StateVector$ is tested. This might undermine the generalizability of the observations as well as the conclusions. In this work, we aim for more universal and comprehensive results. To this end, for each method, we perform $100$ Monte-Carlo simulations with the true relative translation $\mathbf{t}$ uniformly sampled on a sphere centered at $\mathbf{0}^{3\times1}$ with a radius $d_0$ and the true relative heading $\theta$ randomly sampled between $[-\pi,\pi)$. Then, in each simulation, the robots' trajectories are generated randomly. Each trajectory consists of $20$ poses with the distance to the origin no larger than $R_{\max}$. Finally, the odometry data and UWB data are generated from the noise-free data by adding Gaussian noises. In this manner, the results should be indicative of all possible trajectories configuration (shape, size and relative transform) that can be generated from a given condition specified by $d_0$ and $R_{\max}$.

\subsubsection{Effect of trajectory configuration} \label{subsubsec:effect_config}

\textcolor{black}{
Fig. \ref{fig:compare_all_d0_and_Rmax} shows the main results. Overall, our proposed QCQP method provides the best performance, especially in more challenging scenarios (large $d_0$ and small $R_{\max}$). The second best results come from either our SDP method or the NLS method. Regardless of initial guess, our QCQP method is the best option in terms of accuracy. With a rough starting point, the NLS method would be the better option when computation resource is limited. However, as a good initial guess can be time-consuming to obtain during real deployment (especially when the robots are far apart) or even unobtainable (when the robots do not start working at the same time, for example), our SDP method would be the more practical option.}

Generally, it can be seen that both $d_0$ and $R_{\max}$ affects the performance of all methods. The main observations are: the larger the $d_0$, the larger the errors, while $R_{\max}$ has the opposite effect; $d_0$ affects the translation error $e_t$ more noticeably than the heading error $e_{\theta}$ while it is the reverse for $R_{\max}$; the performance of the methods are more similar when $d_0$ is smaller or $R_{\max}$ is larger.

In relation to our theoretical findings: the combination of $d_0$ and $R_{\max}$ would limit the maximum relative angles between successive measurements (which then limits $\alpha$ and $\beta$), while $R_{\max}$ would limit the maximum displacement of the target robot on the horizontal plane (which then limits $\rho$). As such, $d_0$ and $R_{\max}$ together affects $T_i$ and consequently the relative translation $\hat{\mathbf{t}}$, while $R_{\max}$ directly changes $\Phi_i$ and consequently the relative heading $\hat{\theta}$. In principle, we have $\sin\alpha_{\max} \simeq D / d_0$. From a practical perspective, a situation with larger $d_0$ or smaller $R_{\max}$ would be more difficult. Either increasing $d_0$ or decreasing $R_{\max}$ would improve the relative translation estimates, while increasing $R_{\max}$ for the target robot would improve the relative heading estimates.

\subsubsection{Effect of noise levels}

All simulations in this evaluation are run with $d_0 = 50\si{m}$ and $R_{\max} = 10\si{m}$ to imitate an outdoor scenario. The UWB noise $\sigma_r$ and odometry noise $\sigma_o$ vary from ground truth  ($\sim 0.001\si{m}$) to GPS ($\sim 1-10\si{m}$) level of accuracy.
Fig. \ref{fig:compare_all_sigmas_vs_both_errors} illustrates the general influence of $\sigma_r$ and $\sigma_o$ on the estimation errors. \textcolor{black}{
Fig. \ref{fig:compare_all_vs_sigr_sigo} shows the detailed results under extreme conditions.} 

\textcolor{black}{Overall, from Fig. \ref{fig:compare_all_sigmas_vs_both_errors} it can be seen that in situations where measurements are very accurate ($\sigma_r < 0.1$ and $\sigma_o < 0.1$), all methods perform similarly. However, when the noise levels are more realistic ($\sigma_r \geq 0.1$ or $\sigma_o \geq 0.1$), our SDP method clearly outperforms all previous methods. Furthermore, $\sigma_r$ and $\sigma_o$ affect the estimation results in a similar manner but at different scales, with $\sigma_o$ having a stronger impact on the general accuracy. It should be noted that the values of $e_{\theta}$ are mostly the same between methods since unlike $\mathbf{t}$, the value of $\theta$ is limited to $[-\pi, \pi)$ and thus $e_{\theta}$ is bounded.}

\textcolor{black}{
Under extremely large noises (Fig. \ref{fig:compare_all_vs_sigr_sigo}), our methods still provide the best results. The proposed QCQP and SDP methods have mostly identical results, which indicates that the relaxation is tight under the tested conditions. 
Nonetheless, one can argue that the reported translation error is too large for practical purposes. As such, we acknowledge that there is still room for further improvement, in particular dealing with very inaccurate odometry data.}

\subsubsection{Drift-correction capability}

\begin{figure}[t]
\centering
	\includegraphics[width=\linewidth]{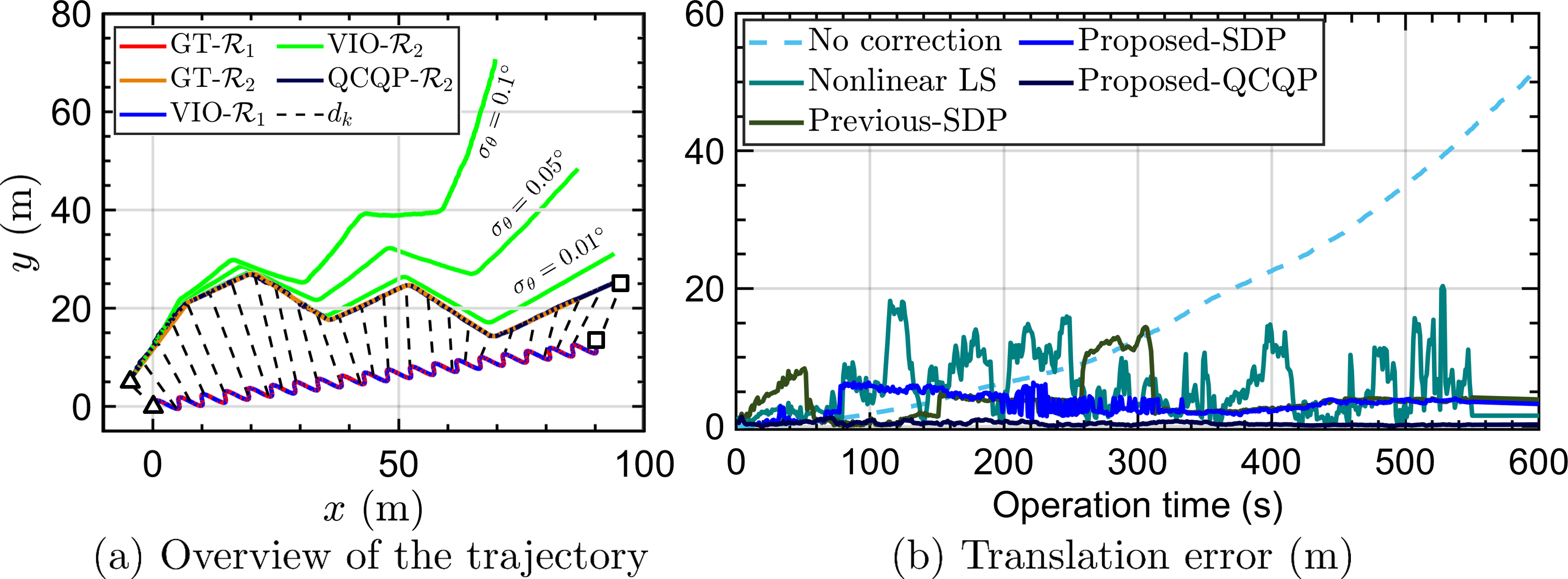}
    \caption{a) The trajectories and UWB measurements. QCQP-$\Robot_2$ refers to the global trajectory of the target robot as computed by our QCQP method, with $\sigma_{\mathbf{t}} = \sigma_{\theta} = 0.1$. Note that only a subset of all $d_k$ is illustrated to improve clarity. b) Translation errors of different methods with $\sigma_{\mathbf{t}} = \sigma_{\theta} = 0.1$.}
    \label{fig:drift_correction_overview}
\end{figure}

While the odometry from SLAM methods can provide accurate short-term ego-motions, they suffer from long-term drift due to the accumulated errors which is particularly prevalent for large-scale missions \cite{shenghai2021ussurvey,nguyen2022viralfusion}. This drift can be modelled as Gaussian noises that acts on the RFT $\prescript{\mathcal{L}_1}{\mathcal{L}_2}{\mathbf{T}}$ \cite{ziegler2021distributed}.
As such, an RTE estimator can monitor and correct the drift by continuously estimating $\prescript{\mathcal{L}_1}{\mathcal{L}_2}{\hat{\mathbf{T}}}_k$ in a sliding window fashion. Let $\sigma_{\mathbf{t}}$ and $\sigma_{\theta}$ be the noise on the translation ($x$, $y$, $z$ axes) and heading $\theta$ parts of $\prescript{\mathcal{L}_1}{\mathcal{L}_2}{\mathbf{T}}$, respectively. The larger $\sigma_{\mathbf{t}}$ and $\sigma_{\theta}$, the more significant the drift (Fig. \ref{fig:drift_correction_overview}a).

\begin{table}[t]
\centering
\begin{adjustbox}{width=\columnwidth}
    \begin{tabular}[t]{
    @{\hskip3pt}c@{\hskip3pt}|
    @{\hskip3pt}c@{\hskip3pt}|
    @{\hskip2pt}c@{\hskip2pt}|
    @{\hskip2pt}c@{\hskip2pt}|
    c|c|c|c}
    \toprule
    $\sigma_{\theta}$
    & \multirow{2}{*}{NC} & Algebra & Linear & NLS & SDP & \multicolumn{2}{c}{Proposed} \\
    \cline{7-8}
    (deg) & 
    & \cite{trawny2010rel3Dtransform} & \cite{molina2019unique} 
    & \cite{ziegler2021distributed} & \cite{jiang2020rel3D} 
    & SDP & QCQP \\
    \hline
    \multicolumn{8}{c}{$\sigma_{\mathbf{t}} = 0.01 \si{m}$} \\ 
    \hline
    $0.01$
    & \underline{2.28} 
    & 84.36 & 94.35 & 6.81 & 4.86
    & 2.36 & \textbf{0.51} \\ 
    \hline
    $0.05$
    & 11.75 
    & 88.06 & 108.83 & 7.59 & 4.91 
    & \underline{2.37} & \textbf{0.42} \\
    \hline
    $0.1$
    & 21.82 
    & 87.86 & 150.11 & 13.96 & 5.02 
    & \underline{2.40} & \textbf{0.58} \\
    \hline
    \multicolumn{8}{c}{$\sigma_{\mathbf{t}} = 0.05 \si{m}$} \\
    \hline
    $0.01$
    & 2.71 
    & 82.07 & 153.26 & 7.05 & 4.94 & 
    \underline{2.37} & \textbf{0.59} \\
    \hline
    $0.05$
    & 11.71 
    & 83.14 & 160.39 & 7.50 & 4.88
    & \underline{2.36} & \textbf{0.53} \\
    \hline
    $0.1$
    & 23.19 
    & 81.16 & 173.13 & 7.06 & 4.97
    & \underline{2.39} & \textbf{0.43} \\
    \hline
    \multicolumn{8}{c}{$\sigma_{\mathbf{t}} = 0.1 \si{m}$} \\
    \hline
    $0.01$
    & 3.48
    & 84.86 & 166.32 & 9.03 & 4.73 & 
    \underline{2.39} & \textbf{0.53} \\
    \hline
    $0.05$
    & 11.09
    & 85.01 & 170.38 & 9.04 & 4.82
    & \underline{2.38} & \textbf{0.54} \\ 
    \hline
    $0.1$
    & 23.06
    & 100.53 & 199.29 & 10.15 & 4.92
    & \underline{2.40} & \textbf{0.61} \\
    \bottomrule
    \end{tabular}
\end{adjustbox}
\caption{RMSE of the translation error ($\si{m}$) of the target robot's aligned trajectory with different drift conditions ($\sigma_{\mathbf{t}}$ and $\sigma_{\theta}$). The \textbf{first} and \underline{second} best results are ranked for each row. NC: no correction.}
\label{table:results_drift_correction}
\end{table}

We simulate a scenario with two robots exploring a large environment (Fig. \ref{fig:drift_correction_overview}a) in a period of $10\si{mins}$. The host robot is equipped with a highly accurate localization system, such as RTK-GPS or LiDAR-based SLAM. The target robot is equipped with a VIO system that will drift away from the ground truth as time goes on. The host robot performs a sinuous trajectory to ensure the observability of the data within the sliding window, while the host robot performs a simple trajectory to scan the area. The sliding window contains the latest $50$ measurements, which are collected in $5$ seconds. 
All measurements are corrupted by $\sigma_o = 0.001\si{m}$ and $\sigma_r = 0.1\si{m}$.
No initial guess was provided, meaning that all methods must 1) estimate the RFT $\prescript{\mathcal{L}_1}{\mathcal{L}_2}{\hat{\mathbf{T}}}_0$ and 2) track the changes of $\prescript{\mathcal{L}_1}{\mathcal{L}_2}{\hat{\mathbf{T}}}_k$, using only recent measurements.

The global position of the target robot in the world frame, i.e. $\prescript{\mathcal{L}_1}{a_2}{\hat{\mathbf{p}}}_k \coloneqq \hat{\mathbf{t}}_k + \hat{\mathbf{C}}_k \prescript{\mathcal{L}_2}{a_2}{\hat{\mathbf{p}}}_k$, is the final output that we are most interested in. Table \ref{table:results_drift_correction} shows the RMSE of the translation error of $\prescript{\mathcal{L}_1}{a_2}{\hat{\mathbf{p}}}_k$. The smaller the RMSE, the better the system in performing both finding and monitoring $\prescript{\mathcal{L}_1}{\mathcal{L}_2}{\hat{\mathbf{T}}}_k$. The NC column shows the original accuracy of the onboard odometry without any correction, which serves as baseline to assess improvements.  Fig. \ref{fig:drift_correction_overview}b demonstrates the translation errors in the case with the most drift.

It is noticeable that the algebraic and linear methods actually worsen the position estimates. The reason is that with only data in a short sliding window, the situation is similar to the “hard" cases described in previous section (small $R_{\max}$, large $d_0$). Hence, these methods could not obtain a good estimate for $\prescript{\mathcal{L}_1}{\mathcal{L}_2}{\hat{\mathbf{T}}}_k$ during the whole operation. The other optimization-based methods show more consistent and improved results in cases with larger drifts. Overall, our QCQP and SDP are the best and second best methods, with a significant enhancement compared to other methods as well as the baseline.


\begin{figure}[t]
\centering
	\includegraphics[width=\linewidth]{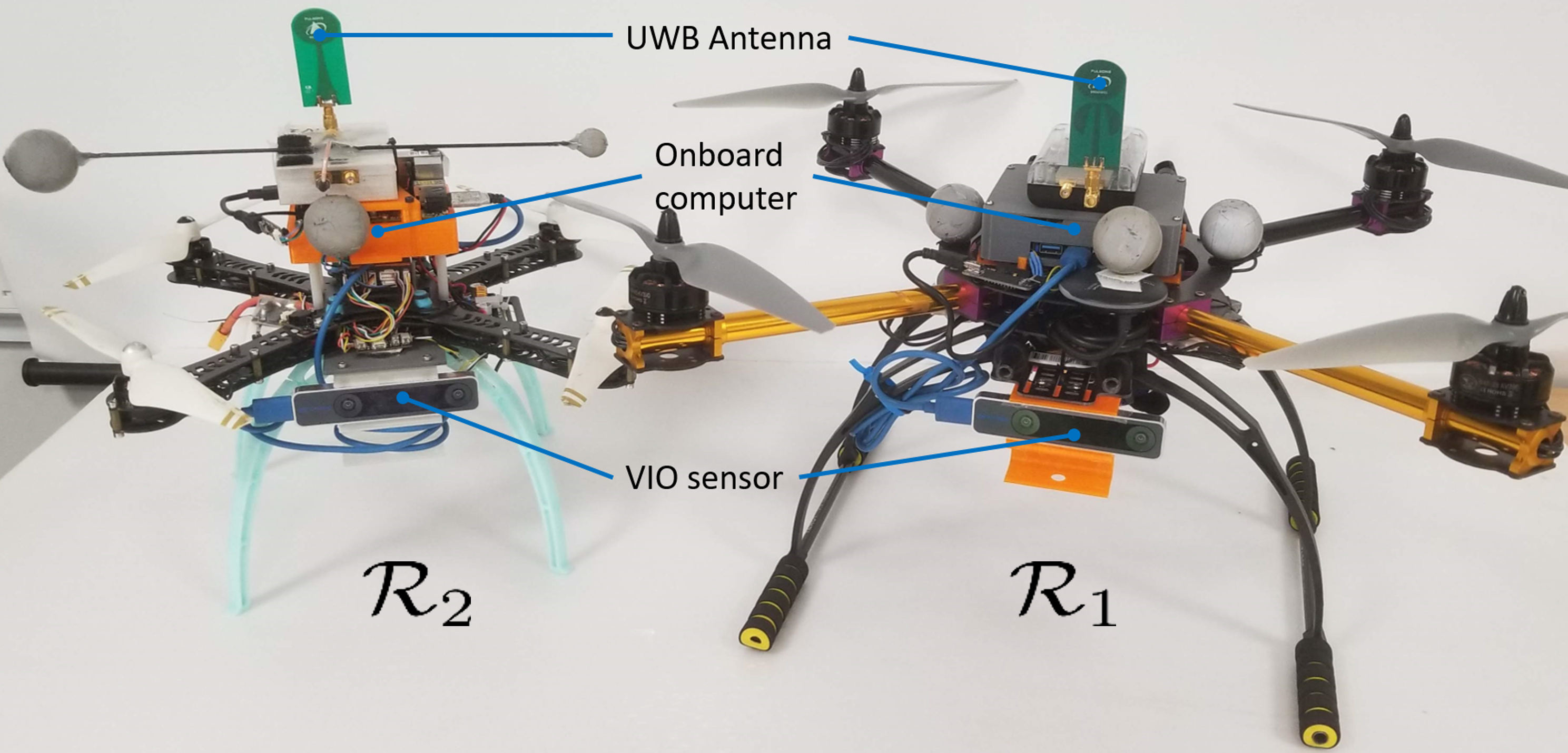}
    \caption{The quadrotor platforms used in our experiments.}
    \label{fig:exp_real_platforms}
\end{figure}

\begin{figure}[t]
\centering
	\includegraphics[width=\linewidth]{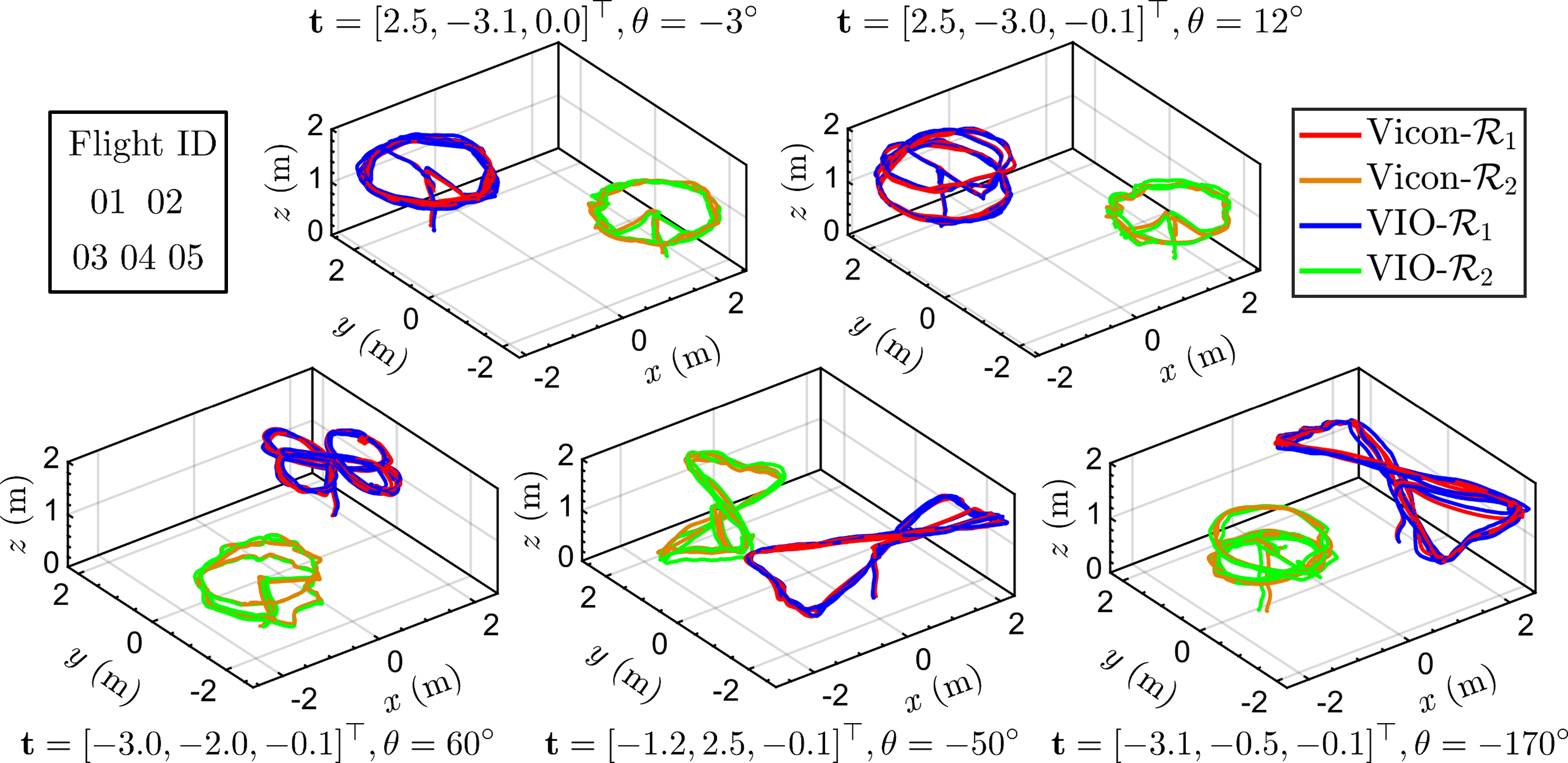}
    \caption{The trajectories in real-life flight tests.}
    \label{fig:exp_real_traj}
\end{figure}

\subsection{Real-life experiments}

\begin{table}[t]
\centering
\begin{adjustbox}{width=\columnwidth}
    \begin{tabular}[t]{
    @{\hskip3pt}c@{\hskip3pt}|
    @{\hskip3pt}c@{\hskip3pt}|
    @{\hskip2pt}c@{\hskip2pt}|
    @{\hskip2pt}c@{\hskip2pt}|
    c|c|c|c}
    \toprule
    \multirow{2}{*}{ID} & \multirow{2}{*}{RMSE} &
        Algebra & Linear & NLS & SDP & \multicolumn{2}{c}{Proposed} \\
    \cline{7-8}
    & &
        \cite{trawny2010rel3Dtransform} & \cite{molina2019unique} &
        \cite{ziegler2021distributed} & \cite{jiang2020rel3D} &
        SDP & QCQP \\
    \hline
    \multirow{2}{*}{01}
    & $\mathbf{t} (\si{m})$ & 
        0.759 & 0.469 & 0.333 & \textbf{0.258} & 0.310 & \underline{0.296} \\
    & ${\theta} (\si{rad})$ & 
        0.057 & 0.311 & 0.053 & \underline{0.047} & \textbf{0.044} & 0.062 \\
    \hline
    \multirow{2}{*}{02}
    & $\mathbf{t} (\si{m})$ & 
        0.384 & 0.456 & 0.295 & 0.379 & \underline{0.203} & \textbf{0.201} \\
    & ${\theta} (\si{rad})$ & 
        0.156 & 0.285 & \textbf{0.072} & 0.102 & \underline{0.082} & \textbf{0.072} \\
    \hline
    \multirow{2}{*}{03}
    & $\mathbf{t} (\si{m})$ & 
        0.570 & 0.339 & 1.855 & \underline{0.130} & 0.131 & \textbf{0.124} \\
    & ${\theta} (\si{rad})$ & 
        0.131 & 0.217 & 0.846 & \underline{0.067} & 0.072 & \textbf{0.059} \\
    \hline
    \multirow{2}{*}{04}
    & $\mathbf{t} (\si{m})$ & 
        0.315 & 0.397 & 0.162 & 0.531 & \underline{0.124} & \textbf{0.108} \\
    & ${\theta} (\si{rad})$ & 
        0.103 & 0.160 & 0.114 & 0.061 & \underline{0.021} & \textbf{0.014} \\
    \hline
    \multirow{2}{*}{05}
    & $\mathbf{t} (\si{m})$ & 
        1.265 & 1.125 & 0.195 & 0.459 & \underline{0.184} & \textbf{0.178} \\
    & ${\theta} (\si{rad})$ & 
        0.211 & 0.483 & 0.132 & 0.138 & \textbf{0.119} & \underline{0.125} \\
    \bottomrule
    \end{tabular}
\end{adjustbox}
\caption{RMSE of translation ($\mathbf{t}$) and heading (${\theta}$) in real-life experiments. Each result is averaged over 03 runs. The \textbf{first} and \underline{second} best results are ranked for each row.}
\label{table:results_real_life}
\end{table}

\begin{figure}[t]
\centering
    \begin{subfigure}[t]{0.49\columnwidth}
    \centering
	\includegraphics[width=\textwidth]{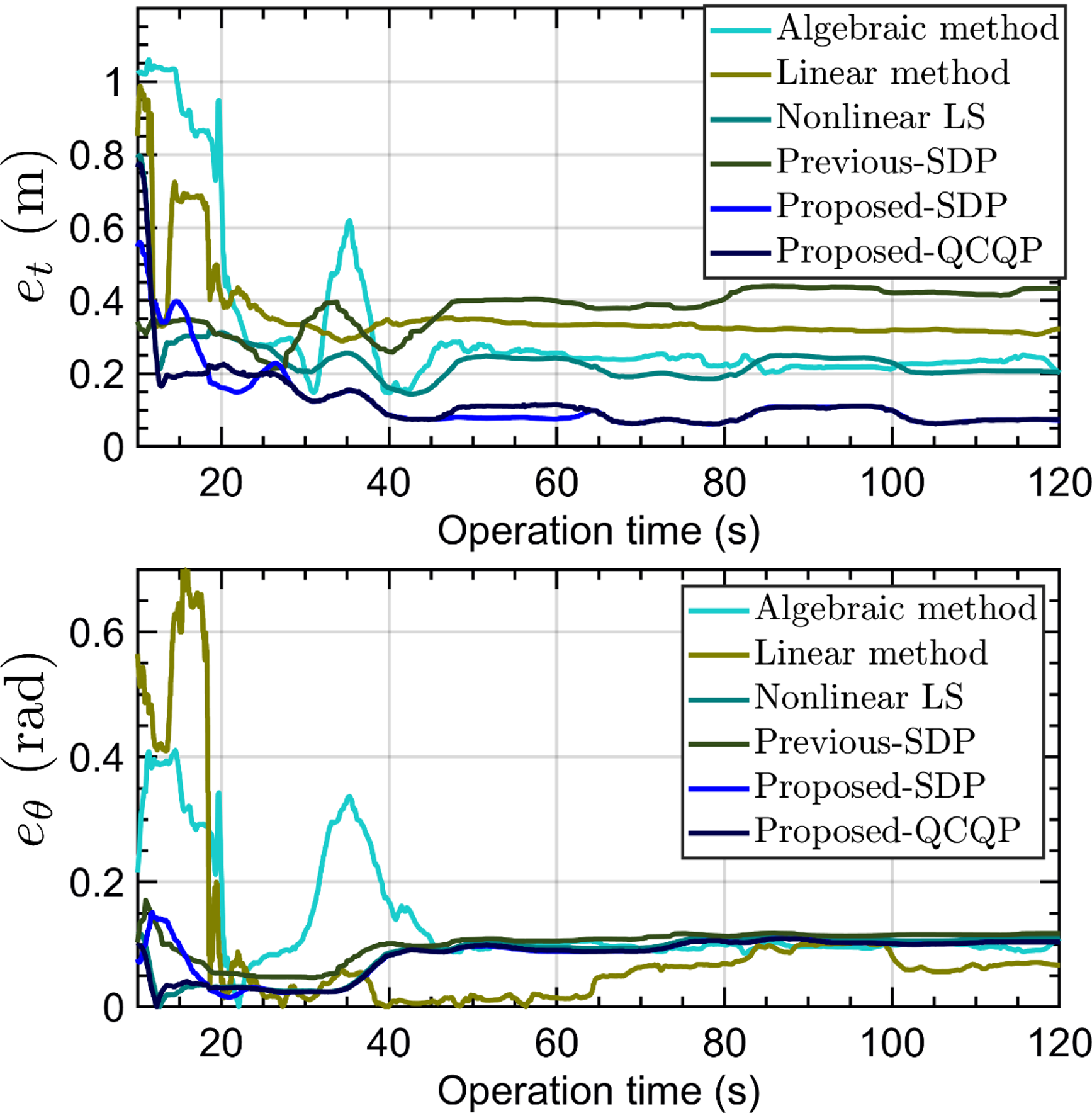}
    \caption{Estimation error}
    \label{fig:flight_02_results_errors}
    \end{subfigure}
    \hfill
    \begin{subfigure}[t]{0.49\columnwidth}
    \centering
	\includegraphics[width=\textwidth]{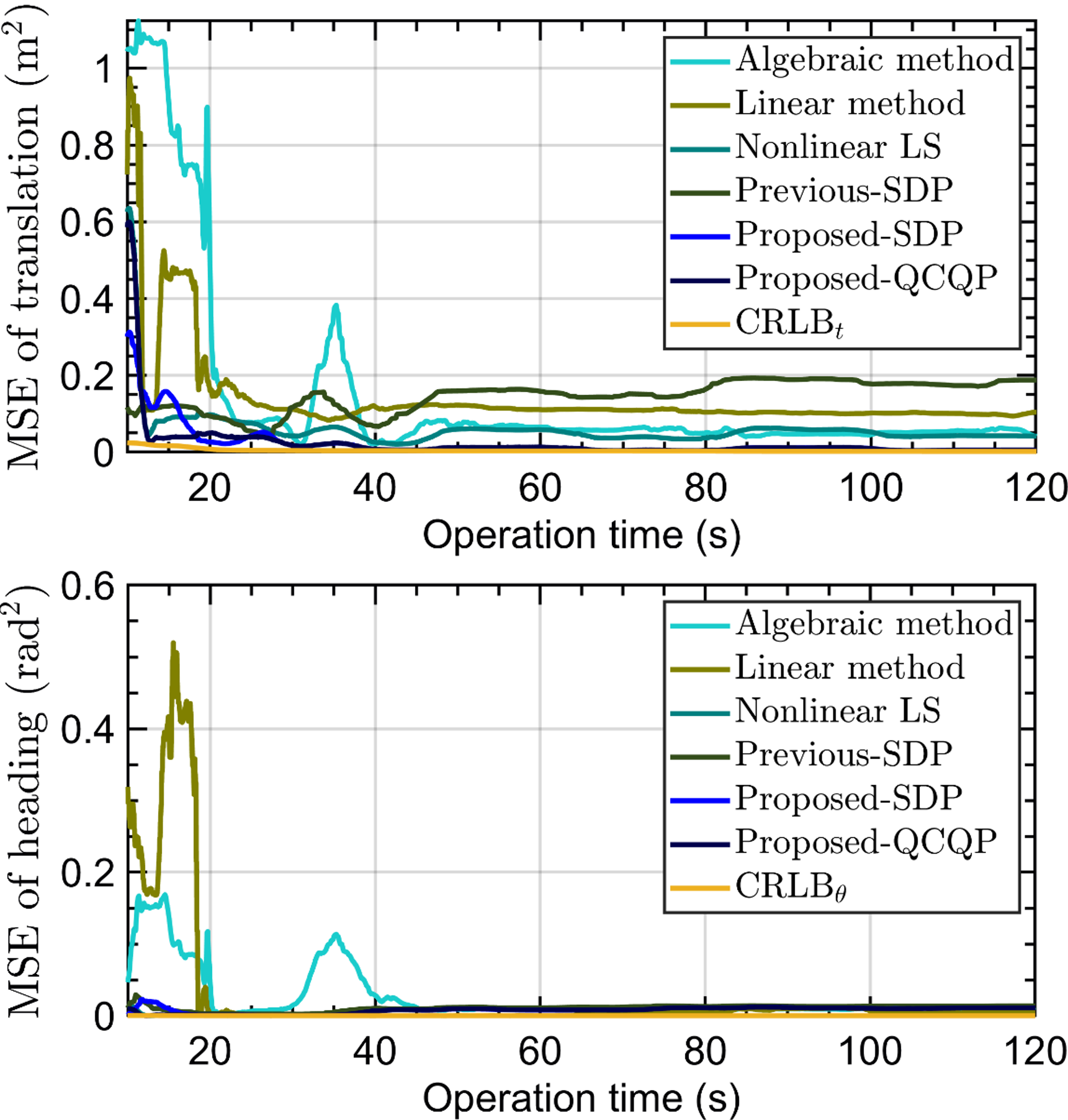}
    \caption{MSE against CRLB}
    \label{fig:flight_02_results_benchmark}
    \end{subfigure}
    \caption{Estimation results in one run for flight 02.}
    \label{fig:flight_02_estimation_results}
\end{figure}

\begin{figure}[t]
\centering
    \begin{subfigure}[t]{0.493\linewidth}
    \centering
    \includegraphics[width=\linewidth]{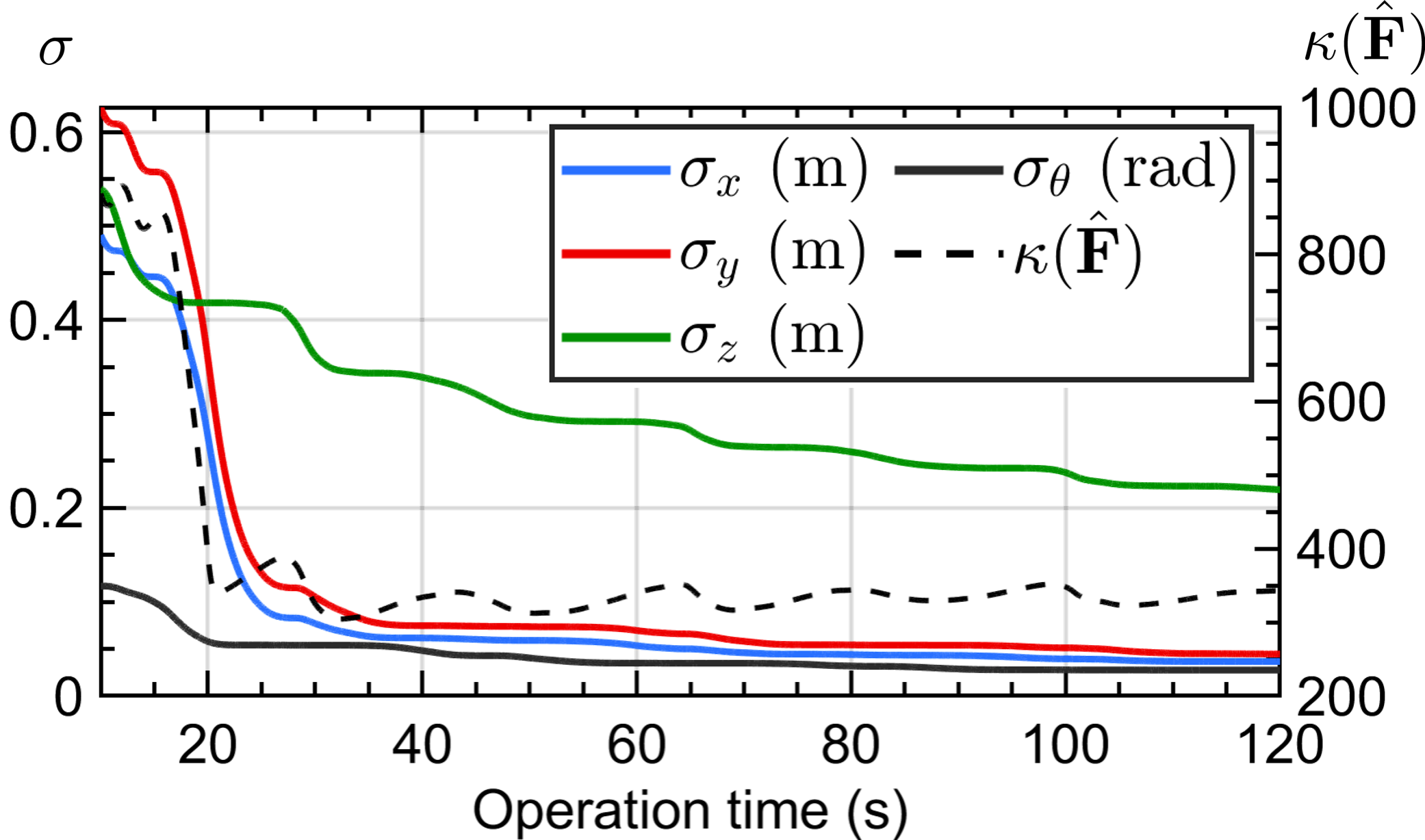}
    \caption{Flight 02.}
    \label{fig:uncertainties_flight_02}
    \end{subfigure}
    \hfill
    \begin{subfigure}[t]{0.493\linewidth}
    \centering
	\includegraphics[width=\linewidth]{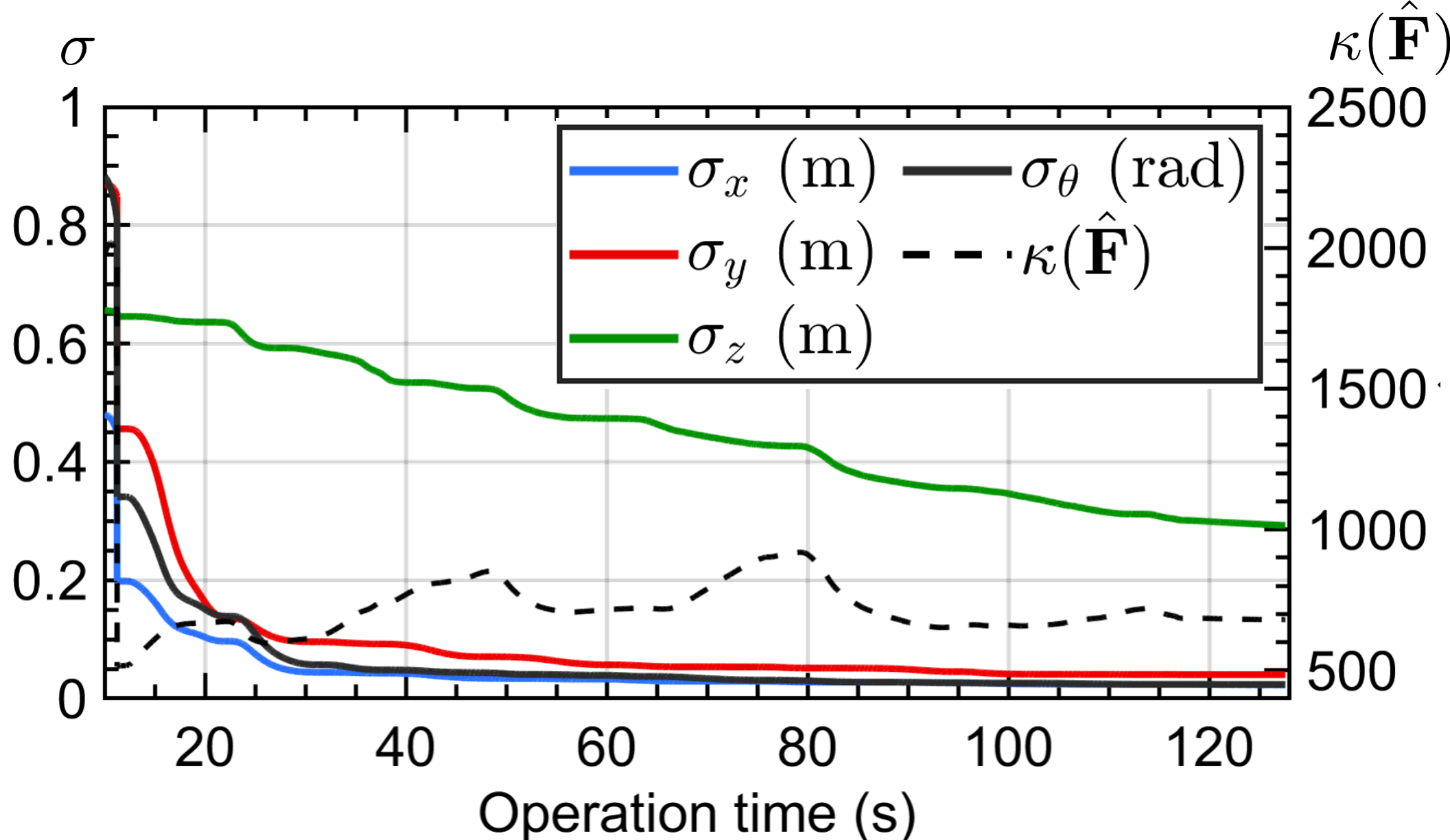}
    \caption{Flight 05.}
    \label{fig:uncertainties_flight_05}
    \end{subfigure}
\caption{Uncertainty of the estimates in two real flight tests.}
\label{fig:uncertainty_metrics_real_flights}
\end{figure}

\begin{figure*}[t]
\centering
    \begin{subfigure}[t]{0.245\textwidth}
    \centering
	\includegraphics[width=\textwidth]{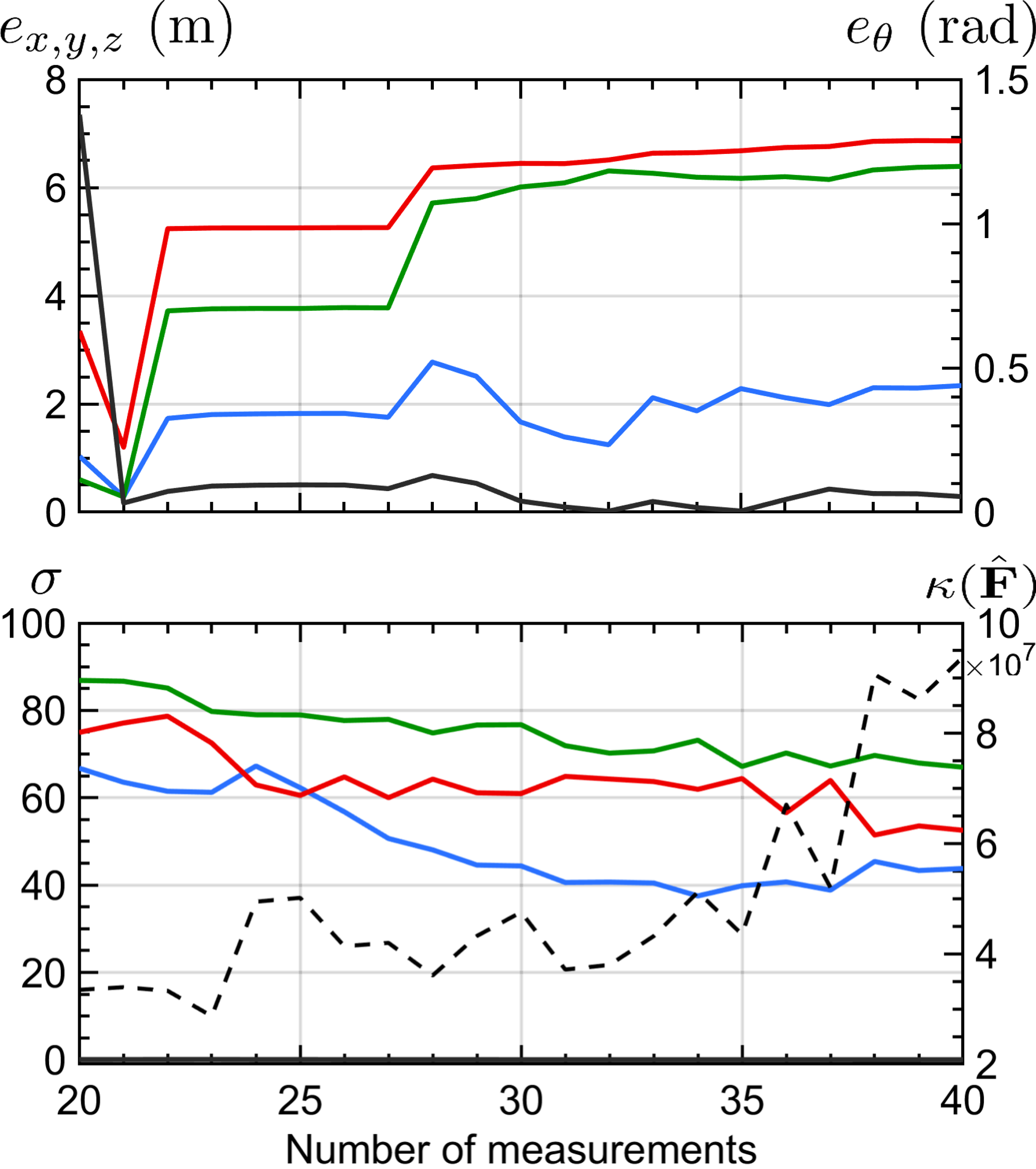}
    \caption{Parallel motion (Fig. \ref{fig:exp_singular_parallel})}
    \label{fig:uncertainties_singular_parallel}
    \end{subfigure}
    \hfill
    \begin{subfigure}[t]{0.245\textwidth}
    \centering
	\includegraphics[width=\textwidth]{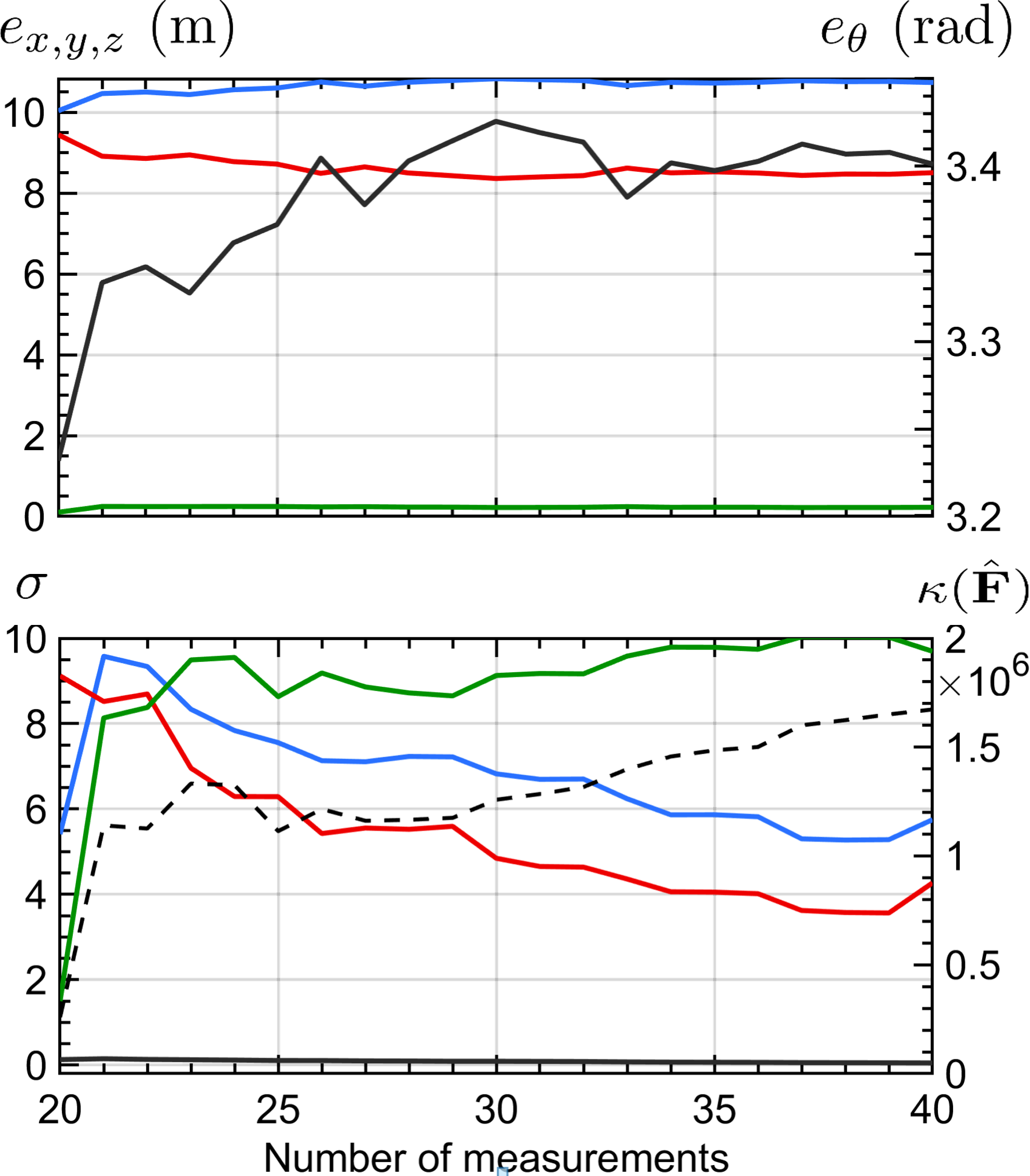}
    \caption{Planar motion (Fig. \ref{fig:exp_singular_planar})}
    \label{fig:uncertaintis_singular_planar}
    \end{subfigure}
    \hfill
    \begin{subfigure}[t]{0.245\textwidth}
    \centering
	\includegraphics[width=\textwidth]{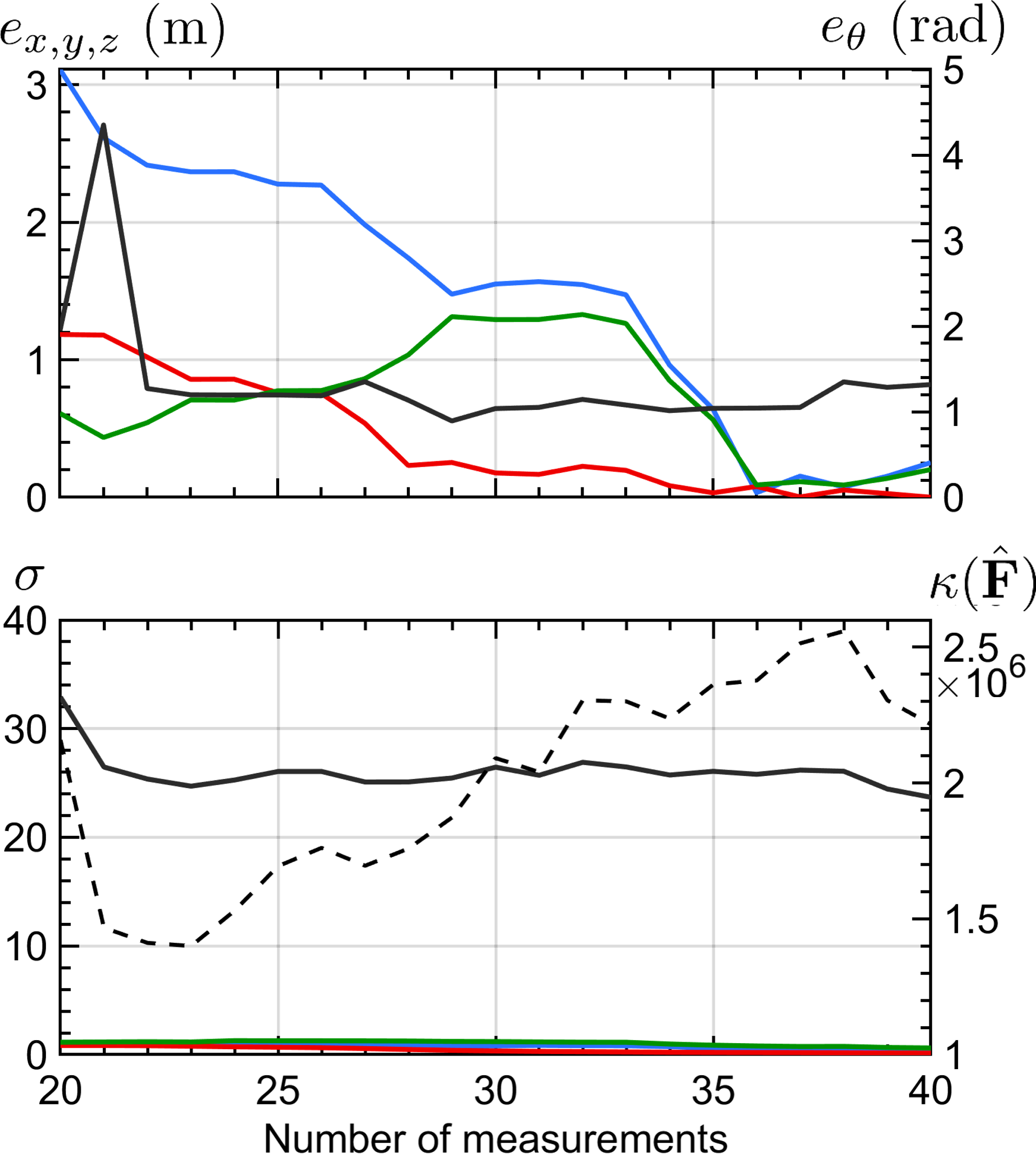}
    \caption{Static target (Fig. \ref{fig:exp_singular_static_target})}
    \label{fig:uncertainties_singular_static_target}
    \end{subfigure}
    \hfill
    \begin{subfigure}[t]{0.245\textwidth}
    \centering
	\includegraphics[width=\textwidth]{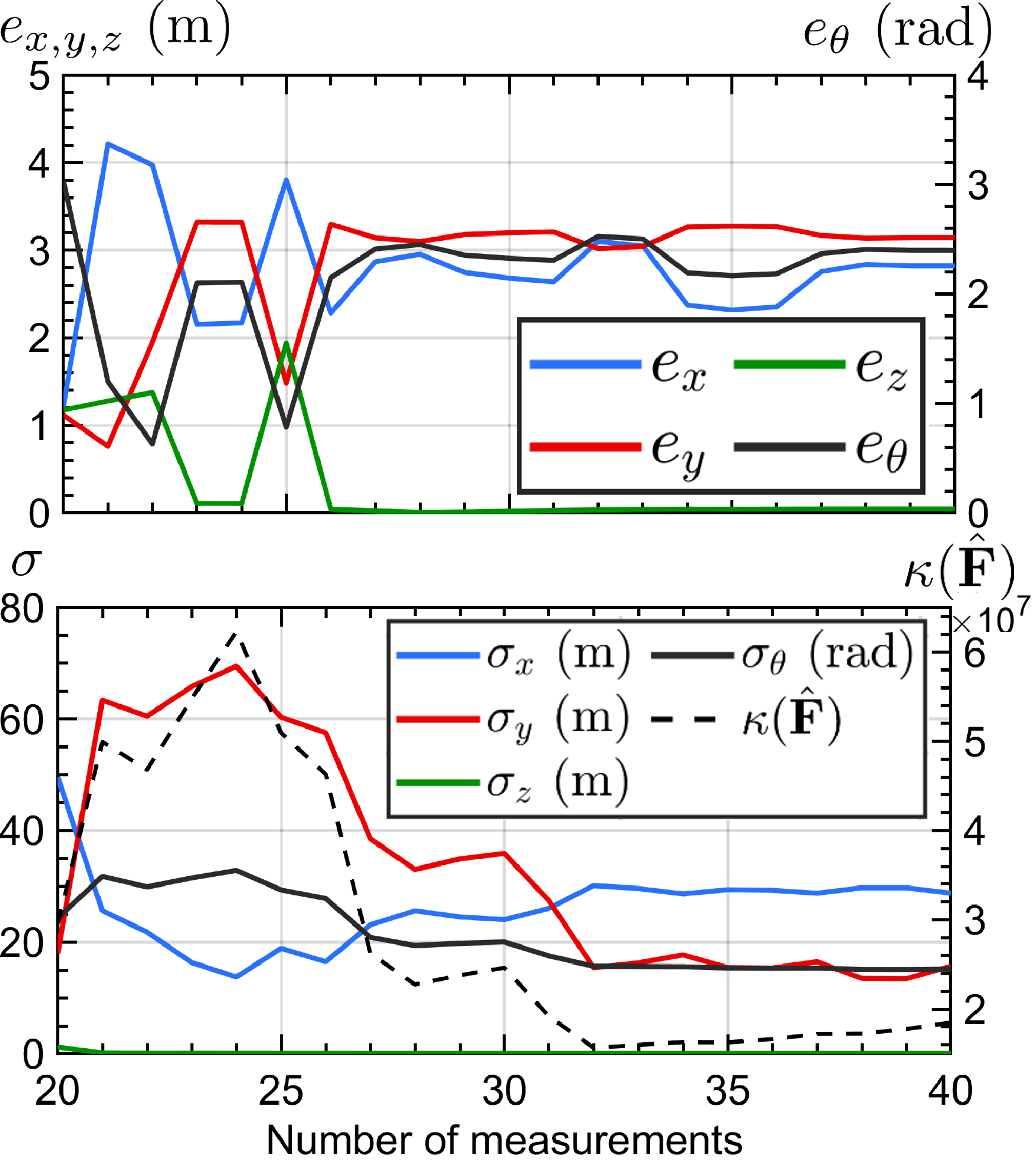}
    \caption{Static host (Fig. \ref{fig:exp_singular_static_host})}
    \label{fig:uncertainties_singular_static_host}
    \end{subfigure}
\caption{Estimation errors (top row) and uncertainty metrics (bottom row) in simulations. We can see that: 1) the estimated uncertainty metrics are clearly much larger in unobservable (Fig. \ref{fig:uncertainties_singular_parallel}-\ref{fig:uncertainties_singular_static_host}) than observable cases (Fig. \ref{fig:uncertainty_metrics_real_flights}), 2) if the standard error $\sigma_i$ or estimation error $e_i$ do not improve over time, the parameter $\Theta_i$ is likely to be unobservable in that configuration.}
\label{fig:uncertainty_metrics_simulations}
\end{figure*}

Fig. \ref{fig:exp_real_platforms} shows the hardware platforms in our real-life experiments. Each quadrotor is equipped with a Humatics P440 UWB\footnote{\url{https://fccid.io/NUF-P440-A}}, an UP2 mini computer\footnote{\url{https://up-board.org/}}, an Intel Realsense T265 VIO sensor\footnote{\url{https://www.intelrealsense.com/tracking-camera-t265/}}. The UWB antenna positions in the body frames are $\prescript{\mathcal{B}_1}{a_1}{\mathbf{p}} {=} [-0.02, 0.1, -0.05]^\top$ and $\prescript{\mathcal{B}_2}{a_2}{\mathbf{p}} {=} [-0.05, 0.15, -0.15]^\top$. The UWB data is generated at $37\si{Hz}$ while the VIO data arrives at $200\si{Hz}$. The noise standard deviations are set as $\sigma_r = 0.1$ and $\sigma_o = 0.002$. We collect the data over 5 flight tests with various trajectory configurations (Fig. \ref{fig:exp_real_traj}) in a VICON room that provides ground truth poses at $100\si{Hz}$. All methods are run on the same Intel NUC i7 mini computer.

Table \ref{table:results_real_life} shows the average results over 3 runs. The algebraic and linear methods get the same level of accuracy for most cases, but the results are not desirable. The NLS and previous SDP method works well only in some scenarios. Our QCQP method surpasses the other methods in most cases while also provides consistent results, with our SDP method being a close second.
Fig. \ref{fig:flight_02_results_errors} illustrates the evolution of the estimates in one run. It is evident that the proposed methods outperform previous approaches.
This result correlates with the performance benchmark in Fig. \ref{fig:flight_02_results_benchmark}, where our methods approach the lower bound faster than others. However, it can be seen that as the operation continues, the estimates can become bias. The reasons might be due to the accumulated drift in the odometry data, errors in the calibration, or any unmodelled effect in the UWB data during the flight.

\subsection{Uncertainty estimation} \label{subsec:UncertaintyEval}

As stated in Sect. \ref{subsubsec:UncertaintyEst}, the uncertainty of the configuration is quantified by the condition number of the estimated FIM, $\condF$, whereas the uncertainty of each parameter is measured by the standard error $\sigma_{\hat{\Theta}_i}$. Fig. \ref{fig:uncertainty_metrics_real_flights} and  \ref{fig:uncertainty_metrics_simulations} demonstrate these values in real-life experiments and simulations, respectively. All simulations are done with $\sigma_r = 0.1$, $\sigma_o = 0.001$, $R_{\max} = 1$ and $d_0 = 3$. All results are obtained from our QCQP method.

In typical observable situations (Fig. \ref{fig:uncertainties_flight_02}, \ref{fig:uncertainties_flight_05}), the estimated uncertainties follow the trend of the actual errors: as more measurements are incorporated, the errors as well as the uncertainties reduce. In simulation, the robots' movements cover all directions and the uncertainty on $x$, $y$, $z$ axes behave similarly. In real flights the motion on the $z$ axis is much more limited than the other axes due to safety concerns and the platforms' capability. As such, the rate of improvement for $\sigma_z$ is noticeably much slower. 
\textcolor{black}{
Noted that for multi-UAV scenarios, the experiments usually start with degenerate configurations: first, the two UAVs take off (i.e., the robots move in parallel in the z axis) then at least one of them will hover for a short period of time to establish communication or collect data (i.e., one robot is static). Hence, by monitoring $\condF$ we can detect whether the configuration is degenerate. As shown in Fig. \ref{fig:uncertainties_flight_02}, the value of $\condF$ stays high until $t=20\si{s}$ which corresponds to when $\Robot_2$ starts moving. Afterwards, a noticeable drop in $\condF$ can be seen which indicates the configuration is no longer unobservable.
}

On the other hand, Fig. \ref{fig:uncertainties_singular_parallel}-\ref{fig:uncertainties_singular_static_host} show the cases where the configuration remains unobservable throughout the experiment. Firstly, notice that the value of the condition number $\condF$ is substantially larger and tend to only increases. Secondly, the particular parameters that are unobservable (shown in Fig. \ref{fig:exp_degenerate_configs}) will have considerably larger standard errors and do not improve over time, while the others behave similar to the observable cases. Hence, these configurations can be recognized as singular with the unobservable DoFs identified.

\subsection{Computational demands}

\begin{figure}[t]
\centering
    \begin{subfigure}[t]{0.95\linewidth}
    \centering
    \includegraphics[width=\linewidth]{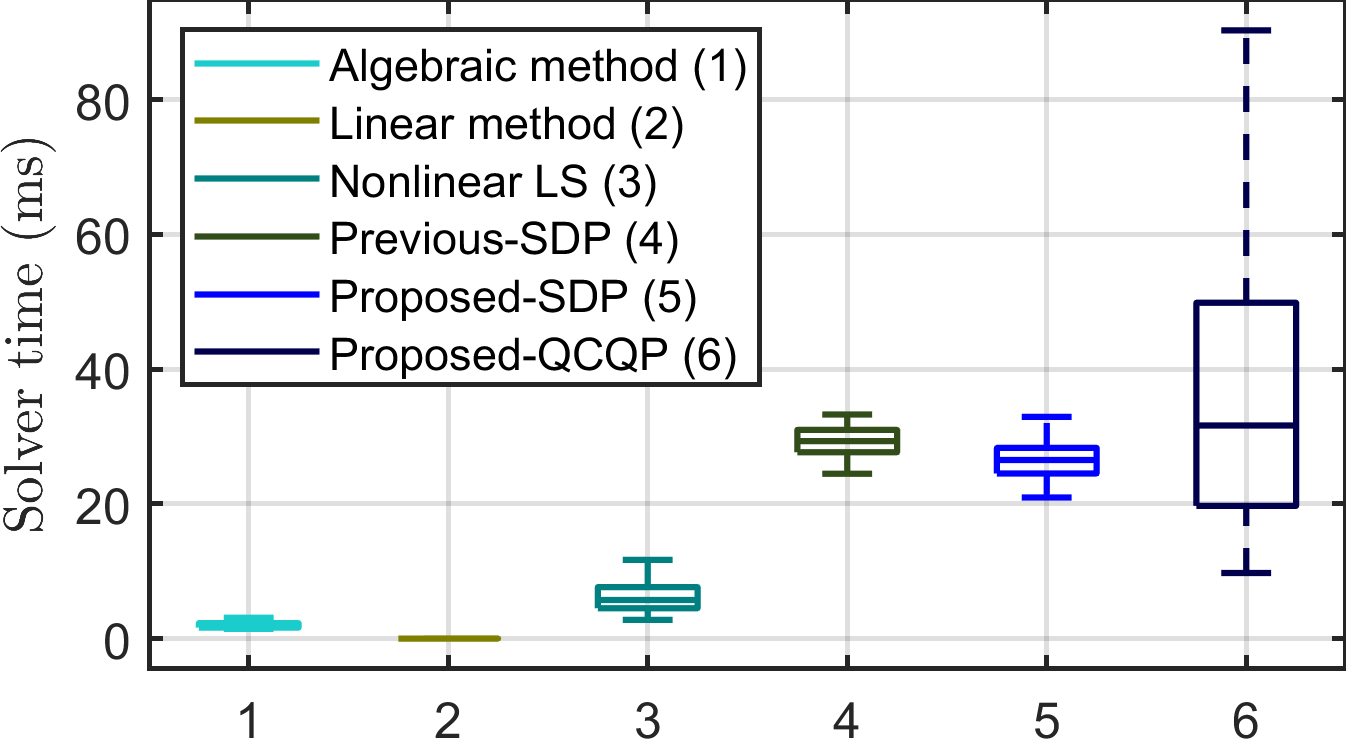}
    \caption{Solver time of all simulations in Fig. \ref{fig:compare_all_d0_and_Rmax}.}
    \label{fig:solver_time_sim_only}
    \end{subfigure}
    \hfill
    \begin{subfigure}[t]{\linewidth}
    \centering
	\includegraphics[width=\linewidth]{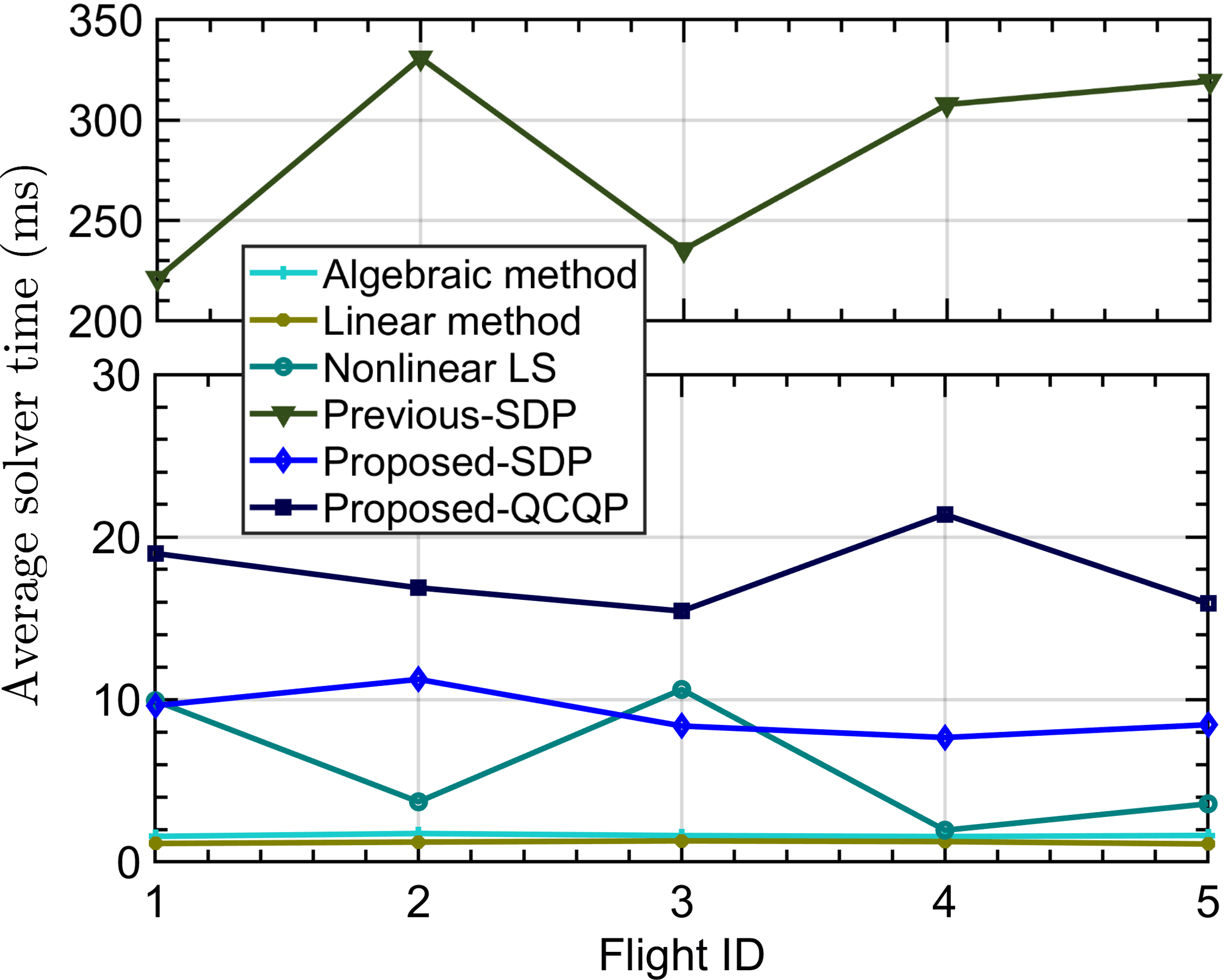}
    \caption{Average solver time in real-life experiments in Table \ref{table:results_real_life}.}
    \label{fig:solver_time_real_only}
    \end{subfigure}
    \caption{Comparison of solver time (ms).}
    \label{fig:solver_time_both_sim_real}
\end{figure}

Fig. \ref{fig:solver_time_sim_only} and \ref{fig:solver_time_real_only} demonstrate the solver time in all simulations of Fig. \ref{fig:compare_all_d0_and_Rmax} and real-life experiments in Table \ref{table:results_real_life}, respectively. Overall, most methods can run in real-time. The linear method is consistently the fastest, closely matched by the algebraic method. Our QCQP method is the slowest in simulation. Since the QCQP method's solver time often correlates with the hardness of the problem (takes longer with smaller $R_{\max}/d_0$ or larger noise) and the simulation covers all cases from easy to hard, the variation is much larger than the other methods.
In real-life experiments, our SDP method runs faster than our QCQP method with an almost $50\%$ improvement, while the previous SDP method is the slowest and not real-time. The reason might be that the QCQP solver library directly supports our problem formulation, while we needed to adapt the public SDP solver library for the formulation of \cite{jiang2020rel3D}. Hence, the optimization of the library would be the main reason for the improved speed of our SDP method. However, as the scale of our real-life experiments is limited, these results are not as indicative as the simulation.

\section{Conclusion} \label{sec:conclusion}

We study the 4-DoF RTE problem using the local odometry and inter-robot UWB range measurements. The theoretical analysis of the problem is put forth, including the CRLB, the FIM and the determinant of the FIM. Based on these findings, insights for the geometric interpretation of information gains for each parameter, methods to detect singular configurations and measure the uncertainty of the estimates are provided. To solve the problem, optimization-based solutions are introduced which consist of a QCQP approach and the corresponding SDP relaxation. Our system outperforms previous methods in both simulations and real-life experiments, especially in challenging scenarios, and are more robust to large UWB noise. While both proposed approaches can run in real-time on mini computers, the QCQP method generally provides the most accurate results but takes longer time than the SDP counterpart. Finding the full unobservable conditions, the optimal trajectories configuration using $\detF$ as well as extending the system to the general case with $N$ robots  are interesting topics for future works.

\section*{Acknowledgements}
We would like to thank Dr. Cao Kun and Mr. Cao Muqing for the fruitful discussions regarding previous works.

\begin{appendices}

\section{}\label{appendix:FIM_full}

After computing the Jacobian, we have
\begin{equation} \label{eq:G_i_original}
\begin{aligned}
    \mathbf{G}_i = 
    \left[
    \partial_x f_i, \partial_y f_i, \partial_z f_i, \partial_{\theta} f_i
    \right] 
    = \frac{1}{d_i} \left[
    g^x_i,\;g^y_i,\;g^z_i,\;g^{\theta}_i
    \right],
\end{aligned}
\end{equation}
where
\begin{equation*}
\begin{aligned}
    d_i 
    &= \norm{ \mathbf{t} + \mathbf{C} \LocPosTarg_i - \LocPosHost_i} 
    = \sqrt{(g^x_i)^2 + (g^y_i)^2 + (g^z_i)^2}, \\
    g^x_i &= g^x_i (t^x, \theta) 
    = t^x + o^x_i \cos{\theta}  - o^y_i \sin{\theta} - \varphi^x_i,\\
    g^y_i &= g^y_i (t^y, \theta) 
    = t^y + o^x_i \sin{\theta} + o^y_i \cos{\theta} - \varphi^y_i,\\
    g^z_i &= g^z_i (t^z) = t^z + o^z_i - \varphi^z_i,\\
    g^{\theta}_i 
    &= g^{\theta}_i (t^x, t^y, \theta) 
    = g^x_i(-o^x_i \sin{\theta}  - o^y_i \cos{\theta})\\ 
    &+ g^y_i(o^x_i \cos{\theta} - o^y_i \sin{\theta}).
\end{aligned}
\end{equation*}

Notice that $g^x_i,g^y_i,g^z_i$ respectively correspond to the displacement in $x,y,z$ axes between the positions of the two robots at time $t_i$. \textcolor{black}{Hence, ${\left[\partial_x f_i, \; \partial_y f_i, \; \partial_z f_i \right]^\top = \frac{1}{d_i}\left[g^x_i,\; g^y_i,\;g^z_i\right]^\top = \mathbf{u}_i}$} where $\mathbf{u}_i$ is a unit vector parallel to the relative position vector $\RelPosUWBAnts_i$ ($\norm{\mathbf{u}_i} = 1$).

We can also simplify $\partial_{\theta} f_i = g^{\theta}_i / d_i$ as
\begin{equation} \label{eq:df_dtheta_simplified}
\begin{aligned}
    &\partial_{\theta} f_i = \frac{1}{d_i} 
    \big[
    g^x_i(-o^x_i \sin{\theta} - o^y_i \cos{\theta})
    + g^y_i(o^x_i \cos{\theta} - o^y_i \sin{\theta})
    \big]\\
    &= 
    \left(
    \begin{bmatrix}
        0 \\
        0 \\
        1
    \end{bmatrix}
    \times 
    \begin{bmatrix}
        o^x_i \cos{\theta} - o^y_i \sin{\theta} \\
        o^x_i \sin{\theta} + o^y_i \cos{\theta} \\
        o^z_i
    \end{bmatrix}
    \right) \cdot
    \begin{bmatrix}
        g^x_i / d_i \\
        g^y_i / d_i \\
        g^z_i / d_i
    \end{bmatrix} \\
    &= 
    \left[ \UnitZ \times  (\mathbf{C} \LocPosTarg_i) \right] 
    \cdot \mathbf{u}_i
    = 
    \left( \UnitZ \times  \AlignedPosTarg_i \right)
    \cdot \mathbf{u}_i
\end{aligned}
\end{equation}
where $\UnitZ = [0,0,1]^\top$. Hence, $\mathbf{G}_i$ can be rewritten as
\begin{equation} \label{eq:G_i_vector}
\begin{aligned}
    \mathbf{G}_i = \left[
    \mathbf{u}_i^\top, \; \Phi_i
    \right],
\end{aligned}
\end{equation}
where $\Phi_i \coloneqq \partial_{\theta} f_i = 
\left( \UnitZ \times  \AlignedPosTarg_i \right) \cdot \mathbf{u}_i$.

\textcolor{black}{
$\Phi_i$ can then be further simplified as
\begin{equation}
\begin{aligned}
    \Phi_i =
    &\left( \UnitZ \times \AlignedPosTarg_i \right)
    \cdot \mathbf{u}_i
    =
    \norm{\UnitZ} 
    \norm{\AlignedPosTarg_i} 
    \norm{\mathbf{u}_i} \\
    &\left| 
        \sin \measuredangle (\UnitZ, \AlignedPosTarg_i)
    \right|
    \cos \measuredangle 
    \left( \UnitZ \times \AlignedPosTarg_i, \mathbf{u}_i \right).\\
\end{aligned}
\end{equation}
Let
$\gamma_i = \frac{\pi}{2} - \measuredangle \left(\UnitZ \times \AlignedPosTarg_i, \mathbf{u}_i\right)$ 
and $\rho_i$ be the length of the projection of $\AlignedPosTarg_1$ on the $xy$ plane of $\{\mathcal{L}_i\}$, we have
\begin{equation*}
\begin{aligned}
    &\sin \gamma_i = \cos \measuredangle 
        \left( \UnitZ \times \AlignedPosTarg_i, \mathbf{u}_i \right),\\
    &\rho_1 \coloneqq
    \norm{\AlignedPosTarg_i}
    \left| 
        \sin \measuredangle (\UnitZ, \AlignedPosTarg_i)
    \right|. \\
\end{aligned}
\end{equation*}
which leads to
\begin{equation} \label{eq:det_Lambda_Phii}
    \Phi_i = \rho_i \sin \gamma_i.
\end{equation}
}

\section{}\label{appendix:sub_problems_derivation}

\subsection{3D RTE with a common heading reference} \label{app_subsec:3D_known_theta}
In this case, the state vector is $\StateVector \coloneqq [t^x, t^y, t^z]^\top$ and the Jacobian is reduced to $\mathbf{G}_i = [\partial_x f_i,\partial_y f_i,\partial_z f_i]$ and we still have the FIM as $\FIM = \sigma_r^{-2} {\mathbf{J}^\top} {\mathbf{J}}$, where the $i$-th row of ${\mathbf{J}}$ is $\mathbf{G}$. Let ${S_1 = \{ 1 \leq i < j < l \leq k \}}$, the Cauchy-Binet formula gives
\begin{equation}
\begin{aligned}
    \det(\FIM) &= \frac{1}{\sigma_r^2} \sum\limits_{S_1}^{}
    \left(\det \left(
    \begin{matrix}
        \mathbf{u}_i^\top \\[0.2em]
        \mathbf{u}_j^\top \\[0.2em]
        \mathbf{u}_l^\top 
    \end{matrix}
    \right) \right)^2 \\
    &= \frac{1}{\sigma_r^2} \sum\limits_{S_1}^{}
    \left(\det \left(
    [\mathbf{u}_i \; \mathbf{u}_j \; \mathbf{u}_l] \right) \right)^2.
\end{aligned}
\end{equation}
Since ${\det([\mathbf{a} \; \mathbf{b} \; \mathbf{c}]) = (\mathbf{a} \times \mathbf{b}) \cdot \mathbf{c}}$, we have
\begin{equation}
\begin{aligned}
    &\det(\FIM) = \frac{1}{\sigma_r^2} \sum\limits_{S_1}^{}
    \left[
    (\mathbf{u}_i \times \mathbf{u}_j) \cdot \mathbf{u}_l
    \right]^2 \\
    &= \frac{1}{\sigma_r^2} \sum\limits_{S_1}^{}
    \norm{\mathbf{u}_i}^2 
    \norm{\mathbf{u}_j}^2 
    \norm{\mathbf{u}_l}^2 
    \sin^2 \alpha \;
    \sin^2 \beta \\
    &= \frac{1}{\sigma_r^2} \sum\limits_{S_1}^{}
    \sin^2 \alpha \;
    \sin^2 \beta,
\end{aligned}
\end{equation}
where 
$\alpha = \measuredangle (\mathbf{u}_i, \mathbf{u}_j)$, 
$\beta = \frac{\pi}{2} - \measuredangle (\mathbf{u}_i \times \mathbf{u}_j, \mathbf{u}_l)$.

\subsection{2D RTE without a common heading reference} \label{subsec:2D_RTE_no_heading}

In this case, the state vector is $\StateVector \coloneqq [t^x, t^y, \theta]^\top$. To simplify the analysis, we still use the same 3D representation for the local odometry vectors and the unit relative position vector $\mathbf{u}_i$ in $\mathbb{R}^3$ but with zero $z$ elements. The target rotation matrix $\mathbf{C} \in SO(3)$ is the same. The length of the projection of $\LocPosTarg_i$ on the $xy$ plane of $\{\mathcal{L}_2\}$ is $\rho_i = \norm{\LocPosTarg_i}$. The range measurement model is
\begin{equation}
\begin{aligned}
    f_i =
    \norm{
    \begin{bmatrix}
        t^x 
        + o^x_i \cos{\theta}  
        - o^y_i \sin{\theta} 
        - \varphi^x_i \\
        t^y 
        + o^x_i \sin{\theta} 
        + o^y_i \cos{\theta} 
        - \varphi^y_i
    \end{bmatrix}
    },
\end{aligned}
\end{equation}
and the Jacobians are $\mathbf{G}_i 
= \left[ \partial_x f_i, \partial_y f_i, \partial_{\theta} f_i \right]
= \frac{1}{d_i} [g^x_i, g^y_i, g^{\theta}_i]$,
where
\begin{equation}\label{eq:hi123}
\begin{aligned}
    d_i &= \sqrt{(g^x_i)^2 + (g^y_i)^2}, \\
    g^x_i &= 
    t^x + o^x_i \cos{\theta} - o^y_i \sin{\theta} - \varphi^x_i, \\
    g^y_i &= 
    t^y + o^x_i \sin{\theta} + o^y_i \cos{\theta} - \varphi^y_i, \\
    g^{\theta}_i &= 
    g^x_i(-o^x_i \sin{\theta}  - o^y_i \cos{\theta}) + g^y_i( o^x_i \cos{\theta} - o^y_i \sin{\theta}).
\end{aligned}
\end{equation} 

The FIM is $\FIM = \sigma_r^{-2} {\mathbf{J}^\top} {\mathbf{J}}$, where the $i$-th row of ${\mathbf{J}}$ is $\mathbf{G}_i$. The Cauchy-Binet formula gives
\begin{equation} \label{eq:det_F2_full}
\begin{aligned}
    &\det(\FIM) = 
    \frac{1}{\sigma_r^2} \sum\limits_{S_2}^{}
    \left(\det \left(
    \begin{bmatrix}
        \partial_x f_{j_1} & \partial_y f_{j_1} & \partial_{\theta} f_{j_1} \\
        \partial_x f_{j_2} & \partial_y f_{j_2} & \partial_{\theta} f_{j_2} \\
        \partial_x f_{j_3} & \partial_y f_{j_3} & \partial_{\theta} f_{j_3} \\
    \end{bmatrix} \right) \right)^2 \\
    &= 
    \frac{1}{\sigma_r^2} \sum\limits_{S_2}^{}
    \left(
    \sum\limits_{i=1}^{3}
    (-1)^{i+3} \; \partial_{\theta} f_l
    \det \left(
    \begin{bmatrix}
        \partial_x f_p & \partial_y f_p \\
        \partial_x f_q & \partial_y f_q 
    \end{bmatrix} \right)
    \right)^2
\end{aligned}
\end{equation}
 where 
 ${S_2 = \{ (j_1, j_2, j_3) \vert 1 \leq j_1 < j_2 < j_3 \leq k \}}$, ${l = j_i}$, ${(p,q) \in \{j_1,j_2,j_3\} \setminus l}$, $p < q$.
From Eq. (\ref{eq:hi123}), we can write $\partial_{\theta} f_l$ as
\begin{equation} \label{eq:2d_no_theta_dtfi}
\begin{aligned}
    \partial_{\theta} f_l  &= 
    \frac{1}{d_l} 
    \Big[
    g^x_l(-o^x_l \sin{\theta} - o^y_l \cos{\theta}) \\
    &+ g^y_l(o^x_l \cos{\theta} - o^y_l \sin{\theta})
    \Big]\\
    &= 
    \left(
    \begin{bmatrix}
        0 \\
        0 \\
        1
    \end{bmatrix}
    \times 
    \begin{bmatrix}
        o^x_l \cos{\theta} - o^y_l \sin{\theta} \\
        o^x_l \sin{\theta} + o^y_l \cos{\theta} \\
        0
    \end{bmatrix}
    \right) \cdot
    \begin{bmatrix}
        g^x_l / d_l \\
        g^y_l / d_l  \\
        0
    \end{bmatrix} \\
    &= 
    \left( \UnitZ \times \AlignedPosTarg_l \right)
    \cdot \mathbf{u}_l
\end{aligned}
\end{equation}
where $\mathbf{u}_l {=} [\frac{g^x_l}{d_l}, \frac{g^y_l}{d_l}, 0]^\top$. Since vector $\AlignedPosTarg_l$ resides on the $xy$ plane, the angle between $\UnitZ$ and $\AlignedPosTarg_l$ is always $\pi / 2$, i.e. $\measuredangle (\UnitZ, \AlignedPosTarg_l) = \pi / 2$. Also, notice that
$\norm{\AlignedPosTarg_l} {=} \norm{\LocPosTarg_l} {=} \rho_l$. Let 
${\gamma_l = \frac{\pi}{2} - \measuredangle 
\left( \UnitZ \times \AlignedPosTarg_l, \mathbf{u}_l \right)}$, 
$\partial_{\theta} f_l$ in Eq. (\ref{eq:2d_no_theta_dtfi}) can be simplified as
\begin{equation} \label{eq:det_2d_dfi_simplified}
\begin{aligned}
    \partial_{\theta} f_l 
    &=
    \norm{\UnitZ}
    \norm{\AlignedPosTarg_l}
    \norm{\mathbf{u}_l}
    \left| 
        \sin \measuredangle (\UnitZ, \AlignedPosTarg_l)
    \right|
    \sin \gamma_l \\
    &= \rho_l \sin \gamma_l.
\end{aligned}
\end{equation}

Next, we have
\begin{equation} \label{eq:det_flfp_2d}
\begin{aligned}
    &\det \left(
    \begin{bmatrix}
        \partial_x f_p & \partial_y f_p \\
        \partial_x f_q & \partial_y f_q 
    \end{bmatrix} \right) 
    = \det \left(
    \begin{bmatrix}
        \partial_x f_p & \partial_x f_q \\
        \partial_y f_p & \partial_y f_q 
    \end{bmatrix} \right) \\
    &= \det \left(
    \begin{bmatrix}
        \partial_x f_p  & \partial_x f_q    & 0 \\
        \partial_y f_p  & \partial_y f_q    & 0 \\
        0                   &                       & 1
    \end{bmatrix} \right) 
    = (\mathbf{u}_p \times \mathbf{u}_q) \cdot \UnitZ.
\end{aligned}
\end{equation}
Since both vectors $\mathbf{u}_p$ and $\mathbf{u}_q$ reside on the $xy$ plane, their cross product will align with $\UnitZ$. The signed angle from $\mathbf{u}_p$ to $\mathbf{u}_q$ can be computed as
\begin{equation}
    \alpha_i = \atantwo 
    \left(
    [\mathbf{u}_p \times \mathbf{u}_q]_z, \;
    \mathbf{u}_p \cdot \mathbf{u}_q
    \right),
\end{equation}
where $[\cdot]_z$ denotes the $z$ element of the argument vector in $\mathbb{R}^3$. As such, 
$\mathbf{u}_p \times \mathbf{u}_q
= \norm{\mathbf{u}_p} \norm{\mathbf{u}_q} \sin\alpha_i \UnitZ 
= \sin\alpha_i \UnitZ$.
Eq. (\ref{eq:det_flfp_2d}) can then be written as
\begin{equation} \label{eq:det_2d_fxfy_simplified}
    \det \left(
    \begin{bmatrix}
        \partial_x f_p & \partial_y f_p \\
        \partial_x f_q & \partial_y f_q 
    \end{bmatrix} \right) 
    = \sin\alpha_i \norm{\UnitZ}^2 
    = \sin\alpha_i.
\end{equation}

Replacing Eq. (\ref{eq:det_2d_dfi_simplified}) and (\ref{eq:det_2d_fxfy_simplified}) into (\ref{eq:det_F2_full}), we have
\begin{equation}
    \det(\FIM) = 
    \frac{1}{\sigma_r^2} \sum\limits_{S_2}^{}
    \left[
    \sum\limits_{i=1}^{3}
    (-1)^{i+1}
    \rho_i
    \sin{\alpha_i}
    \sin{\gamma_i}
    \right]^2.
\end{equation}

\subsection{2D RTE with a common heading reference}
In this case, the state vector is $\StateVector \coloneqq [t^x, t^y]^\top$. Built upon the notations and equations in Appendix \ref{app_subsec:3D_known_theta}, the determinant of the FIM is
\begin{equation}
\begin{aligned}
    &\det(\FIM) = 
    \frac{1}{\sigma_r^2} \sum\limits_{S_3}^{}
    \left(\det \left(
    \begin{bmatrix}
        \partial_x f_i & \partial_y f_i \\
        \partial_x f_j & \partial_y f_j \\
    \end{bmatrix} \right) \right)^2 \\
    &= 
    \frac{1}{\sigma_r^2} \sum\limits_{S_3}^{}
    \left(
    (\mathbf{u}_i \times \mathbf{u}_j) \cdot \UnitZ
    \right)^2
    = 
    \frac{1}{\sigma_r^2} \sum\limits_{S_3}^{}
    \sin^2\alpha,
\end{aligned}
\end{equation}
where $S_3 = \{(i,j) \; \vert \; 1 \leq i < j \leq k \}$ and $\alpha = \measuredangle(\mathbf{u}_i,\mathbf{u}_j)$.
This completes the proof.

\end{appendices}

\balance
\bibliographystyle{IEEEtran}
\bibliography{IEEEabrv,./references}

\end{document}